\documentclass[final,12pt]{colt2025} % Anonymized 
\title[Online Experimental Design Under Network Interference]{Online Experimental Design With Estimation-Regret Trade-off Under Network Interference}

\usepackage{blindtext}

\newcommand{\A}{{\mathcal{A}}}
\newcommand{\U}{{\mathcal{U}}}

\newcommand{\bH}{{\mathcal{H}}}
\newcommand{\bP}{{\mathcal{P}}}
\newcommand{\bR}{{\mathcal{R}}}
\newcommand{\bS}{{\mathbf{S}}}
\newcommand{\bbH}{{\mathbb{H}}}
\newcommand{\bC}{{\mathcal{C}}}
\newcommand{\bbS}{{\mathcal{S}}}
\newcommand{\bO}{{\mathcal{O}}}
\newcommand{\bK}{{\mathcal{K}}}
\newcommand{\bbE}{{\mathcal{E}}}
\newcommand{\bone}{{\bold{1}}}
\newcommand{\N}{{\mathcal{N}}}
\newcommand{\T}{{\mathcal{T}}}

\usepackage{makecell}
\usepackage{multirow} 
\usepackage{extarrows}

\usepackage{diagbox}
\usepackage{booktabs}
\usepackage{multirow}
\usepackage{lipsum}
\usepackage{bbm}
\usepackage{algorithm}
\usepackage{algorithmic}

\usepackage{graphicx}
\usepackage{subcaption}

\usepackage{bbm}
\usepackage{float}%提供float浮动环境
\usepackage{booktabs}%提供命令\toprule、\midrule、\bottomrule

\usepackage{amsfonts}
\usepackage{amsmath}
\usepackage{bm}
\usepackage{amssymb}
\usepackage{graphicx} 
\usepackage{nccmath}
\usepackage{thmtools}
\usepackage{thm-restate}
\usepackage{hyperref}
\usepackage{cleveref}
\usepackage{color}
\usepackage{cite}
\usepackage{natbib}
\usepackage{tikz}

\newtheorem{condition}{Condition}
\DeclareMathOperator*{\argmax}{arg\,max}

\newcommand{\cC}{\mathcal{C}}

\newcommand{\cO}{\mathcal{O}}

\newcommand{\bE}{\mathbb{E}}
\usepackage{lastpage}
\firstpageno{1}

\begin{document}

\coltauthor{%
 \Name{Zhiheng Zhang$^*$} \Email{zhiheng-20@mails.tsinghua.edu.cn}\\
 \addr Institute for Interdisciplinary Information Sciences, Tsinghua University
 \AND
 \Name{Zichen Wang$^*$} \Email{zichenw6@illinois.edu}\\
 \addr Coordinate Science Laboratory$\And$Computer Engineering, University of Illinois Urbana-Champaign
 % \AND
 % \Name{Yuhao Wang} \Email{yuhaow@tsinghua.edu.cn}\\
 % \addr Institute for Interdisciplinary Information Sciences, Tsinghua University \\ Shanghai Artificial Intelligence Laboratory and Shanghai Qi Zhi Institute
}

% \centering{Preliminary draft}

\maketitle
\def\thefootnote{*}\footnotetext{These authors contributed equally to this work. This is the first version of this preliminary draft submitted by Zhiheng Zhang (correspondence email: \texttt{zhiheng-20@mails.tsinghua.edu.cn}). We sincerely welcome all kinds of suggestions.}\def\thefootnote{\arabic{footnote}}

% \def\thefootnote{$\dagger$}\footnotetext{This is a preliminary draft.}\def\thefootnote{\arabic{footnote}}

% {\begin{center}
%     Preliminary draft{$^\dagger$}
% \end{center}
% }

~\quad\\
% {Preliminary draft\footnote{This is a preliminary draft uploaded from Zhiheng Zhang to assist in applying for a postdoctoral position.}}
% \center{}
% \center{}
% \center{}

% \\~\quad~\\
% \begin{abstract}%   <- trailing '%' for backward compatibility of .sty file
% Network interference reflects diverse sociological behaviors, wherein the treatment assigned to one individual within a network may influence the outcome for other individuals, such as their neighbors. It has garnered significant interest in both the field of causal inference and bandit. However, the objective of the two fields are very different: the causal inference community favors estimators of treatment effect with high accuracy, while the online learning community seeks to develop sequential policy for minimizing the regret. Hence, designing an online policy that simultaneously ensures statistical power and regret reduction is of interest. In this paper, we develop a unified interference-based stochastic bandit framework. Compared to existing literature, we expand the definition of arm space by leveraging the statistical concept of exposure mapping. We establish the Pareto-optimal trade-off between the regret and estimation accuracy with respect to time period and arm space, which also remains superior to the baseline in the absence of network interference. Correspondingly, we propose an algorithmic implementation and model generalization. 
% % Our objective is to bridge the statistics and learning communities by employing network interference as a facilitating mechanism.
% \end{abstract}

\begin{abstract}
    Network interference has attracted significant attention in the field of causal inference, encapsulating various sociological behaviors where the treatment assigned to one individual within a network
may affect the outcomes of others, such as their neighbors. A key challenge in this setting is that
standard causal inference methods often assume independent treatment effects among individuals,
which may not hold in networked environments. To estimate interference-aware causal effects, a
traditional approach is to inherit the independent settings, where practitioners randomly assign experimental participants into different groups and compare their outcomes. While effective in offline
settings, this strategy becomes problematic in sequential experiments, where suboptimal decision
persists, leading to substantial regret. To address this issue, we introduce a unified interference-
aware framework for online experimental design. Compared to existing studies, we extend the
definition of arm space by utilizing the statistical concept of exposure mapping, which allows for
a more flexible and context-aware representation of treatment effects in networked settings. Cru-
cially, we establish a Pareto-optimal trade-off between estimation accuracy and regret under the
network concerning both time period and arm space, which remains superior to baseline models
even without network interference. Furthermore, we propose an algorithmic implementation and
discuss its generalization across different learning settings and network topology.
\end{abstract}


\section{Introduction}
% Network interference has received significant attention in both the fields of causal inference~\citep{leung2022causal, leung2022rate, leung2023network} and online statistical learning theory~\citep{agarwal2024multi, jia2024multi}. Unlike the SUTVA assumption~\citep{imbens2024causal}, network interference considers a more general scenario where the treatment received by one individual may influence the outcomes of others. Network interference model has been widely used in economics~\citep{arpino2016assessing, munro2021treatment} and social sciences~\citep{bandiera2009social, bond201261, paluck2016changing, imbens2024causal}. 
Network interference has attracted significant attention in the fields of causal inference~\citep{leung2022causal, leung2022rate, leung2023network} and online statistical learning theory~\citep{agarwal2024multi, jia2024multi}, due to its capability to capture more complex real-world interactions. Unlike the SUTVA assumption~\citep{imbens2024causal}, which posits that the treatment assignment and outcomes are isolated to individuals, network interference acknowledges the influences that treatments received by one individual may have on the outcomes of others within a network. This model has found extensive application in economics~\citep{arpino2016assessing, munro2021treatment} and social sciences~\citep{bandiera2009social, bond201261, paluck2016changing, imbens2024causal}, where understanding such interconnected dynamics is crucial.

To successfully identify causal effect under network interference, one straightforward way is to conduct randomized experiments and use the difference in means type estimators to estimate causal effect based on the experimental data~\citep{leung2022causal, leung2022rate, leung2023network,gao2023causal}. Such design is related to many applications~\citep{ciotti2020covid, cai2015social}. For instance, \citet{ciotti2020covid} suggested the randomized experiment on a group of volunteering patients to investigate the therapeutic average treatment effects of various drugs for influenza, e.g., COVID-19, where each individual's status of cure is influenced by the treatment assignment of their neighboring individuals. In practice, an experiment may consist of multiple rounds, and researchers may wish to use the experimental data from the previous rounds to enhance the social welfare of the experimental participants by minimizing the regret of the future rounds~\citep{mok2021managing}. This requires us to consider the trade-off between the \emph{estimation accuracy} of the causal effect and the \emph{cumulative regret} of the experiment. Apparently, such an online experiment represents a more complex design than offline. For example, if experimental designers directly borrow the Bernoulli sampling in offline design~\citep{leung2022causal}, they would empirically result in a regret linear to round time due to the lack of optimal strategy exploration. This motivates us to design a sequential policy that theoretically guarantees the optimal trade-off between the two objectives under interference. Besides, such sequential policy is also relevant to multi-armed bandits with network interference literature~\citep{jia2024multi, agarwal2024multi}, which focuses primarily on minimizing regret rather than improving estimation accuracy.

To reiterate, it is crucial to recognize that estimation efficiency and regret might not be optimized simultaneously, necessitating a careful consideration of the trade-off between these two objectives. Optimal estimation efficiency, such as the Bernoulli design above, generally requires that the sampling probability of each arm remains strictly greater than zero, where the sub-optimal decision persists, leading to substantial regret. Conversely, optimal algorithms, such as the Upper Confidence Bound (UCB)~\citep{Auer2002FinitetimeAO} and its variants, employ probability-vanishing exploration strategies for sub-optimal arms, potentially violating the overlap assumption in causal inference~\citep{d2021overlap}. This violation limits the estimator's precision, as the overlap assumption is critical for ensuring valid causal inferences by maintaining sufficient data across all arms~\citet{sekhon2009opiates}.

Existing works that explore the estimation-regret trade-off often overlook the presence of network interference, effectively assuming a scenario where only a single individual is considered throughout the experiment. Perspectives include empirical algorithm design~\citep{liang2023experimental}, theoretical bi-objective optimization~\citep{simchi2023multi}, and analyses of the interaction between trade-offs and exogenous model assumptions~\citep{duan2024regret}. In comparison, our work extends such a trade-off in the context of network interference. Integrating the aforementioned perspectives requires an elevated viewpoint to construct a challenging yet more universally applicable framework. Specifically, we introduce a unified online network interference-based experimental design setting, referred to as \texttt{MAB-N}. This setting extends the definition of arm space in the multi-armed bandit (MAB) literature by employing the statistical concept of exposure mapping~\citep{leung2022causal, aronow2017estimating}. We derive the theoretical optimal estimation-regret trade-off within it and provide an algorithmic implementation capable of achieving this optimal balance. Our contributions are summarized as follows:

% In this trade-off scenario, some initial work has now appeared to explore the trade-off between these two goals~\citep{liang2023experimental, simchi2023multi}; however, this work has remained on independent samples without interference. 
% To our knowledge, we provide the first theoretical trade-off result between these two interests on the network. 

% \subsection{Literature review}

% \paragraph{Overview of our results}
\begin{itemize}
    \item We establish a unified setting for online experimental design with network interference, referred to as \texttt{MAB-N}, which leverages the statistical concept of exposure mapping.
    
    \item We bridge the multi-objective minimax trade-off, achieving Pareto-optimality between treatment effect estimation and regret efficiency under network interference. Additionally, we propose criteria for a MAB algorithm to achieve Pareto-optimality.
 
    \item We propose the \texttt{UCB-TSN} algorithm to achieve the aforementioned Pareto trade-off by constructing an upper bound for both the ATE estimation error and regret, which is also validated by experiments. Our \texttt{UCB-TSN} algorithm outperforms the elegant preliminary work in (i) the degenerated single-unit case without interference and (ii) the extended adversarial bandit setting.

\end{itemize}

Our paper is organized as follows: Section~\ref{related} provides a brief literature review. Section \ref{framework} introduces our general \texttt{MAB-N} setting and discusses Pareto-optimality to illustrate the estimation-regret trade-off. Section \ref{pareto} provides a general lower bound for the joint performance of regret and estimation, followed by the criteria for any algorithm to achieve Pareto optimality. Section \ref{alg} proposes the Pareto-optimal algorithmic implementation and includes a comparison with the baseline. Section \ref{adversarial_sec} extends \texttt{MAB-N} to adversarial cases. Finally, Section~\ref{discussion} concludes the paper with further discussion.

\section{Related Work}\label{related}

% Our results primarily bridge two lines of research: (i) the extension of bandit modeling scenarios by incorporating interference settings from the statistical community~\citep{agarwal2024multi,jia2024multi}. (ii) the trade-off between estimation and regret in online learning (without network interference)~\citet{simchi2023multi,duan2024regret}, as seen in Table~\ref{tab_compare} in Appendix~\ref{app_review}. For the first line, \citet{agarwal2024multi} utilizes Fourier analysis to reformulate interference-aware bandit as a sparse linear stochastic bandit. However, they only considered interference between first-order neighbors and impose a sparsity assumption to limit the number of neighbors each node can have. In contrast, \citet{jia2024multi} study bandit under interference without this assumption but require a switchback design, forcing all nodes to adopt the same arm simultaneously, which overlooks cases where the optimal arm varies across nodes or subgroups. For the second line, to the best of the authors' knowledge, \citet{simchi2023multi} was the first to establish a rigorous trade-off between regret and estimation error. Additionally, \citet{duan2024regret} argue that this Pareto-optimality can be further improved—specifically, both regret and estimation error can simultaneously reach their optimal values—under the additional assumption of so-called covariate diversity. We direct the reader to the Appendix~\ref{app_review} for further details on the related work.

Our results primarily bridge two lines of research: (i) extending bandit modeling scenarios by integrating interference settings from the statistical community~\citep{agarwal2024multi,jia2024multi}, and (ii) exploring the trade-off between estimation and regret in online learning without network interference~\citep{simchi2023multi,duan2024regret}, as detailed in Table~\ref{tab_compare} in Appendix~\ref{app_review}. In the first line of research, the insightful work of \citet{agarwal2024multi} creatively utilizes Fourier analysis to reformulate interference-aware bandits as sparse linear stochastic bandits. This innovative approach, however, focuses on interference among first-order neighbors and incorporates a sparsity assumption to limit the number of neighbors each node can have. Complementing this, the meticulous study by \citet{jia2024multi} advances the understanding of bandits under interference by forgoing such assumptions, though their methodology requires a switchback design. This design insists that all nodes adopt the same arm synchronously, potentially overlooking scenarios where the optimal arm varies across nodes or subgroups. Turning to the second line of research, we commend \citet{simchi2023multi} for pioneering a rigorous trade-off between regret and estimation error. Additionally, \citet{duan2024regret} contribute significantly by proposing enhancements to this Pareto-optimality, suggesting that both regret and estimation error might simultaneously reach their optimal levels under the thoughtful assumption of covariate diversity. We invite readers to explore further details on these related works in Appendix~\ref{app_review}.

\section{Framework}\label{framework}

\paragraph{Classic MAB under network interference.}
We introduce our setting following~\citet{agarwal2024multi}, which generalizes~\citet{Auer2002FinitetimeAO,simchi2023multi} to the network interference. We focus on the stochastic bandit problem involving a $K$-armed set $\mathcal{K} = \{k\}_{k = 0}^{K-1}$, an $N$-unit set $\mathcal{U} = \{i\}^N_{i = 1}$, and the time horizon $t \in [T]$. The relationship between units is encoded in the adjacency matrix $\mathbb{H}:= \{h_{ij}\}_{i,j \in \U}$\footnote{It does not mean we must get all information about $\mathbb{H}$; instead, it depends on our detailed design.}, where \( h_{i,j} = 1 \) signifies that units \( i \) and \( j \) are neighbors, whereas \( h_{i,j} = 0 \) otherwise. $K,N,\mathbb{H}$ are predefined. At each round, unit interactions induce interference effects. The \textit{original super arm} is represented by an $N$-dimension vector $A_t := (a_{1,t},...,a_{N,t}) \in \mathcal{K}^\U$. To bridge this formulation to causal inference, we start by notating the so-called potential outcome in statistics~\citep{rubin2005causal} (expected reward in the bandit community~\citep{Auer2002FinitetimeAO}) as $\{Y_i(A_t)\}_{i \in \U} = \{Y_i(a_{1,t}, a_{2,t},...a_{N,t})\}_{i\in \U}$ for unit $i$ in time $t$\footnote{Unit $i$'s potential outcome is only related to the treatments of the total population via a fixed function, as is standard in interference-based causality~\citep{leung2022causal, leung2022rate, leung2023network}. This setting relaxes the traditional “Stable Unit Treatment Value Assumption” (SUTVA)~\citep{rubin1980randomization}, which assumes that one unit's outcome is unaffected by others' treatments.}. Without loss of generality, we set $\forall i \in \U$, $A\in\bK^\U$, $Y_i(A) \in [0,1]$. In this sense, the \textit{single-unit reward} of unit $i$ upon time $t$ is given by $r_{i,t}(A_t) = Y_i(A_t) + \eta_{i,t}$, where $r_{i,t}(.)$ represents the reward function of unit $i\in \U$, and $\eta_{i,t}$ is zero-mean i.i.d. 1-sub Gaussian noise for each unit. Finally, we define instance $\nu$ as any legitimate choice of $\{\mathcal{D}(Y_i(A))\}_{i\in\U,A\in\bK^\U}$, where $\mathcal{D}(Y_i(A))$ denotes the reward distribution of unit $i$ if super arm $A$ is pulled; and then denote $\mathcal{E}_{0}$ as the set of all feasible $\nu$. Our primary interest is designing a learning policy $\pi:= (\pi_1,...,\pi_T)$. In round $t$, the agent observes the history $\bH_{t-1} = \big\{A_1,\{r_{i,1}(A_1)\}_{i\in \U},...,A_{t-1},\{r_{i,t-1}(A_{t-1})\}_{i\in \U}\big\}$, where each term is an $N$-dimensional vector. The policy $\pi_t$ is a probabilistic map from $\bH_{t-1}$ to the next action $A_t$. We denote $\pi_t(A) = \mathbb{P}_{\pi}(A_t = A \mid \bH_{t-1})$ indicating the probability that a super arm $A$ is selected in round $t$.

\paragraph{Additional notation.} 
% We define $\bm{e}_i$ as a vector, where the $i$-th element is $1$ and all other elements are 0. We write $[Q]:= \{1,2,\dots,Q\}$ for any $Q \in \mathbb{N}^+$. We define $a \vee b:= \max\{a,b\}$ and $a \wedge b := \min\{a,b\}$. For sequences of all positive numbers $\{a_n\}_{n \in \mathbb{N}^+}$ and $\{b_n\}_{n \in \mathbb{N}^+}$, we denote $a_n = O(b_n)$ if $\exists C>0$, such that $\forall n, a_n \leq C b_n$; to the contrary, we denote $a_n = \Omega(b_n)$ if $\exists C>0$, such that $\forall n, a_n \geq C b_n$. Moreover, we denote $a_n = \Theta(b_n)$ if $a_n = O(b_n)$ and $a_n = \Omega(b_n)$ both hold. We also denote $a_n = \Tilde{O}(b_n)$ if $\exists C>0, k \in \mathbb{N}^+ \cup \{0\} $, such that $a_n \leq C b_n (log(b_n))^k$.

We define \(\bm{e}_i\) as the standard basis vector whose \(i\)-th element is \(1\) and all other elements are \(0\). For any \(Q \in \mathbb{N}^+\), we use the shorthand notation \([Q] := \{1,2,\dots,Q\}\). We define the operations:
$
a \vee b := \max\{a,b\}, \quad a \wedge b := \min\{a,b\}.
$  For sequences of positive numbers \(\{a_n\}_{n \in \mathbb{N}^+}\) and \(\{b_n\}_{n \in \mathbb{N}^+}\), we adopt the following asymptotic notations: \(a_n = O(b_n)\) if there exists a constant \(C > 0\) such that for all sufficiently large \(n\),  
  $
  a_n \leq C b_n.
  $; \(a_n = \Omega(b_n)\) if there exists a constant \(C > 0\) such that for all sufficiently large \(n\),  
  $
  a_n \geq C b_n.
  $; \(a_n = \Theta(b_n)\) if both \(a_n = O(b_n)\) and \(a_n = \Omega(b_n)\) hold. Finally, \(a_n = \Tilde{O}(b_n)\) if there exist constants \(C > 0\) and \(k \in \mathbb{N}^+ \cup \{0\}\) such that  
  $
  a_n \leq C b_n (\log b_n)^k.
  $

% \zhiheng{1 Ue; remove the assumption; do not need instance-dependent result; 2 we highlight $\mathcal{U}_{\mathcal{E}}$ is important..}

% \zhiheng{exposure mapping only treats the informative part, but not the noise part (independent!!)}

\subsection{Motivation: the hardness of classic MAB under interference}

In this framework, referring to the concept of {cumulative regret} in traditional MAB problems~\citep{lattimore2020bandit}, the performance metric of policy $\pi$ could be identified as
\begin{equation}
\begin{aligned}
\bR^{{naive}}(T, \pi) := \frac{T}{N} \sum_{i \in \U} 
 Y_{i}(A^*) - \mathbb{E}_{\pi} \Bigg[ \frac{1}{N} \sum_{t  \in [T]} \sum_{i \in \U} r_{i,t}(A_t) \Bigg] , ~A^* := \arg\max_{A \in \mathcal{K}^\U} \frac{1}{N}\sum_{i\in\U} Y_i(A).
\end{aligned}
\end{equation}
Foreseeably, a fundamental challenge in this setting is that the original super arm suffers from an exponentially large action space ($|\mathcal{K}^{\mathcal{U}}| = {K}^N$), making direct optimization infeasible. Given this computational burden, we first establish a \textit{negative result} to illustrate that directly pursuing the policy $\pi$ using the original super arm is computational \textit{impractical}.

% \zhiheng{$K^N$ could be generalized to be the action space size...for the interference, it is particularity important...}

% \paragraph{Negative result:} 
\begin{proposition}\label{negative}
      Given a priori $N, K, \mathbb{H}$. For any policy $\pi$, there exists a hard instance $\nu \in \mathcal{E}_0$ such that $ \bR^{\text{naive}}_\nu(T, \pi) =   \Omega \big( \frac{1}{\sqrt{N}} (T \wedge \sqrt{K^NT})\big)$. 
\end{proposition}
% When $N=1$, it degenerates to the traditional researchers' consensus of the lower bound analysis in MAB, which is simply written as $\Omega(\sqrt{KT})$~\citep{lattimore2020bandit} with $T \geq K$. However, For any pre-fixed $\{N,K\}$ in the network interference, Proposition~\ref{negative} exhibits a more non-trivial negative result. 
% Proposition~\ref{negative} indicates that the regret convergence rate is dominated by the relative size between time period and arm space, and thus exhibits as a two-piece function: when $T \leq K^N$ under interference, regret $\bR^{naive}_\nu(T, \pi)$ grows linearly with $T$; to the contrary, when $T \geq K^N$, although it degenerates to a square root rate with respect to $T$, unfortunately, it is additionally harmed by an {exponentially} large parameter $\sqrt{{K^N}/{N}}$. This predicament also verifies why~\citet{agarwal2024multi} should additionally consider the interference only from first-order neighbors and introduce sparsity assumptions, otherwise it would not be possible to obtain a meaningful regret.
Proposition~\ref{negative} reveals that the regret convergence rate is influenced by the relative size of the time period compared to the arm space, resulting in a two-piece function. Specifically, when $T \leq K^N$ under interference, the regret $\bR^{naive}_\nu(T, \pi)$ increases linearly with $T$. Conversely, otherwise, although the rate degenerates to a square root relative to $T$, it is adversely affected by an exponentially large parameter $(\sqrt{{K^N}/{N}})$. This negative result, from a counter perspective, substantiates why~\citet{agarwal2024multi} and~\citet{jia2024multi} respectively relaxed the model from the network topology and action space: \citet{agarwal2024multi} prudently considers interference only from first-order neighbors and incorporates sparsity assumptions, while~\citet{jia2024multi} restrict the action space to the all one and all zero $N$-dimensional vector. Without such considerations, obtaining meaningful regret bounds would be unfeasible.

Further, it manifests more insights upon the triple of concepts (i) time, (ii) regret, and (iii) arm space, than lower bound analysis in classic MAB~\citep{lattimore2020bandit}. It is because researchers tend to preemptively judge that ``$\text{time period} \gg \text{arm numbers}$'', e.g., force $N=1$ in the single-unit setting and then $T \gg K$ holds by default. However, this oversimplification consideration of arm space can be detrimental under the interference scenario. For instance, even if we just choose $K=2, N=30$,  any algorithm under interference-based {MAB} setting would potentially be cursed by an impractical regret. In sum, these insights motivate us to develop a general statistical framework to allow for a more reasonable reduction in the action space dimension without imposing excessive assumptions on the network topology, which is the so-called \texttt{MAB-N}, illustrated as follows.
% The centric part is constructing a reasonable performance metric for the above learning policy, ...
% In such a sense, we first introduce \textit{exposure mapping} for dimension reduction to prepare for the regret construction. 

\subsection{Setting: \texttt{MAB-N}} 
% \zhiheng{I will give two figures following Example 5 to show different exposure mapping definitions that could represent the ``direct effect'' and ``spillover effect'', respectively.}...
We introduce the concept of \textit{exposure mapping} developed by~\citet{leung2022causal, aronow2017estimating}. We define the pre-specified function mapping from the original super arm space ($\mathcal{K}^\mathcal{N}$) to a $d_s$-cardinality discrete values ($d_s \ll K^N$) taking advantage of the network structure.
% \footnote{It does not lose generalization, since any mm-dimensional ll-cardinality vector could be represented as an mlm^l-cardinality scalar.}.
For clarity, we consider the discrete function case:
\begin{equation}
s_{i} := \bS(i, A, \mathbb{H}), \text{~where~}
\boldsymbol{S}: \mathcal{U} \times \mathcal{K}^{\U} \times \mathbb{H} \rightarrow \U_s,~ |\U_s| = d_s. \label{exposure_def}
\end{equation}
Here $\mathcal{U}_{s}$ is called as {exposure arm set}. We set $S = \{\textbf{S}(i,A,\mathbb{H})\}_{i \in \U} \equiv (s_{1},\dots,s_{N})$ as the \textit{exposure super arm}, and then we can decompose the policy $\pi_t(\cdot)$ and define the exposure-based reward:
\begin{equation}
\begin{aligned}
 &{\pi_t(A)} := { \mathbb{P}_{} (A_t = A \mid \bH_{t-1}) } =  {\mathbb{P}_{}(A_t = A \mid S_t) } {{\mathbb{P}_{}}({S_t} \mid \bH_{t-1})},\\
&[\Tilde{Y}_i({S}_{t}), \Tilde{r}_{i,t}({S}_{t})]^{\top} := \sum_{A \in \mathcal{K}^{\U}} [Y_i(A), {r}_{i,t}(A)]^{\top} \mathbb{P}_{}(A_t =A \mid {S}_t), 
\end{aligned}\label{exposure}
\end{equation}

The second line of Eq~\eqref{exposure} generalizes the framework of~\citet{leung2022causal} by incorporating a broader class of exposure mappings. Specifically, while the original formulation assumes a fixed exposure structure, our approach allows for a more flexible characterization of treatment assignments under network interference. Detailed derivations are deferred to Appendix~\ref{app_exposure_sec}.  To formalize in practice, we could define \( \mathbb{P}(A_t = A \mid S) \) as a \textit{predefined}, \textit{time-invariant} sampling rule, which the learner specifies before the learning process begins. For example, in the case of uniform sampling (by default), we have: $
 \mathbb{P}_{\pi}(A_t = A \mid S) = {\sum_{A\in\mathcal{K}^\U} \delta\{ \mathbb{A}\}}/|\mathbb{A}|, $
 where \( \delta(\cdot) \) is an indicator function, and \( \mathbb{A}:= \{A: \{\textbf{S}(i, A,\mathbb{H})\}_{i \in \U} = {S}\} \) denotes the set of all assignments that result in the observed exposure state \( S_t \). This formulation ensures that if \( S \) does not match the set \( \{\textbf{S}(i,A,\mathbb{H})\}_{i \in \mathcal{U}} \), the probability of selecting \( A_t = A \) given \( S \) is zero. Conversely, if \( S \) corresponds to this set, then \( A \) is chosen with strictly positive probability, i.e., \( \mathbb{P}(A_t = A \mid S) > 0 \). Under this framework, the observed outcome \( \Tilde{Y}_i(S_{t}) \) in Eq~\eqref{exposure} depends solely on the network topology \( \mathbb{H} \) and the exposure state \( S_t \), independent of the specific arm assignment \( A_t \). This highlights a key property of exposure mapping: it abstracts away individual-level treatment assignments while preserving the structural dependencies induced by network interference. To further quantify decision-making performance under network interference, we introduce the exposure reward \( \Tilde{r}_{i,t}({S}_{t}) \), which serves as a proxy for the expected reward in the exposure space\footnote{Notably, the difference between \( \Tilde{Y}_{i}(S_{t}) \) and the empirically observed reward \( {r}_{i,t}(A_t) \) arises from two distinct noise components: (i) sampling noise, where practitioners approximate \( \Tilde{r}_{i,t}(S_t) \) using samples of \( {r}_{i,t}(A_t) \), and (ii) endogenous noise, inherited from the original variability \( \eta_{i,t} \) in the observed reward. A detailed discussion on noise rescaling is provided in Appendix~\ref{app_exposure_sec}.}. Building on this exposure-based representation, we now define the regret function, which quantifies the performance gap between the optimal and chosen policies under exposure mapping.

% \zhiheng{delete the citation of David...}
% \zhiheng{revise the pi-A-S definition..it is fixed}

\paragraph{Regret based on exposure mapping.} According to the action space reduction in Eq~\eqref{exposure}, we provide a more general and realistic regret compared to~\citet{jia2024multi, simchi2023multi,agarwal2024multi} (refer to Example~$1$-$4$).
% \begin{equation}
% \begin{aligned}
% \bR(T, \pi) = \frac{1}{N}\sum_{t \in [T]} \sum_{i \in [N]} \bigg( \Tilde{r}_{i,t}(s^*) - \Tilde{r}_{i,t}(\textbf{S}(i,A_t,\mathbb{H}))) \bigg), 
% \end{aligned}\label{regret}
% \end{equation}
% where s∗=\argmaxs∈\Us∑i∈[N]\TildeYi(s)s^* = \argmax_{s\in \U_s} \sum_{i \in [N] } \Tilde{Y}_i(s). 
% However, the simultaneous exposure arm for all units might not always be achievable (via the original arm within action space K\U\mathcal{K}^{\U}). Moreover, nodes in different clusters might prefer specific exposure arms.\zhiheng{For instance...} Hence, we propose a generalized version via clustering under mild assumptions.
We define the clustering set $\bC := \{\mathcal{C}_q\}_{q \in [C]}, C = |\mathcal{C}|$ where
$\forall i\neq j, i,j \in [C], \mathcal{C}_i \cap \mathcal{C}_j = \varnothing, \cup \{\mathcal{C}_q\}_{q \in [C]} = \U $. For brevity, we denote $\mathcal{C}^{-1}(i)$ as the cluster of node $i$. We define the exposure-based regret:
\begin{equation}
\begin{aligned}
\bR_{\nu}(T, \pi) =  \frac{T}{N} 
 \sum_{i \in \U} 
 \Tilde{Y}_{i}(S^* ) - \frac{1}{N} \mathbb{E}_{\pi} \Bigg[\sum_{t \in [T]} \sum_{i \in \U} \Tilde{r}_{i,t}(S_t) \Bigg],~S^* = \argmax_{S\in \U_\mathcal{E}  } \sum_{i \in \U } \Tilde{Y}_i(S  ),
\end{aligned}\label{regret_cluster}
\end{equation}
where exposure arm space $\U_\bbE := \U_\bC \cap \U_\bO$ with $\U_{\bC} := \big\{S: \forall i,j \in \U, \mathcal{C}^{-1}(i) = \mathcal{C}^{-1}(j) \text{~implies}~S \bm{e}_i= S \bm{e}_j \big\}$ and $\mathcal{U}_{\mathcal{O}} := \big\{\{\textbf{S}(i, A, \mathbb{H})\}_{i \in \U}:  A \in \mathcal{K}^{\mathcal{U}} \big\}$. Here, $\U_\mathcal{C}$ denotes all kinds of ideally cluster-wise switchback exposure super arm. For instance, if $\mathcal{U}_s \in \{0,1\}, N=4, \mathcal{C}_1 = \{1,2\}, \mathcal{C}_2 = \{3,4\}$, then $\mathcal{U}_{\mathcal{C}} = \{(k_1,k_1, k_2,k_2):{k_1,k_2 \in \{0,1\}}\}$. Moreover, $\mathcal{U}_{\mathcal{O}}$ includes all exposure arm sets compatible with the original arm set. It induces that $\vert \U_\mathcal{E} \vert \le |d_s|^C$. Essentially, during the exposure mapping process, we efficiently reduce the action space by condensing the original arm information in a structured manner, thereby achieving a controlled enhancement of regret efficiency. According to Proposition~\ref{negative}, this balance between sacrifice and gain emerges naturally and inevitably. Such cluster-wise exposure mapping structures have appeared in multiple prior works. We illustrate how our framework can surrogate previous settings as special cases. By assigning specific parameter values, we can (i) flexibly transition between these cases (the following examples), (ii) allow for an adaptive balance in different scenarios (Table~\ref{compare} in Appendix~\ref{app_review}), and (iii) even characterize new and more general real-world scenarios (experiments in Appendix~\ref{ex_app}) where existing methods would fundamentally fail. 

\paragraph{Comparison with previous literature.} For the comparison of regret, \textbf{Example (i)} Classic MAB ~\citep{Auer2002FinitetimeAO,simchi2023multi} considered the case $N=1$, i.e., single unit without network, and $\boldsymbol{S}(1, A, \mathbb{H}) := A$, $A\in\mathcal{K}$.
% Moreover, current interference-based MAB literature~\citep{agarwal2024multi, jia2024multi} are both included as our special instances:
\textbf{Example (ii)} 
    \citet{agarwal2024multi} chooses $\boldsymbol{S}(i, A, \mathbb{H}):= A\bm{e}_i$ and $C=N$ (each unit is assigned to a separate cluster). \textbf{Example (iii)} On the other hand,~\citet{jia2024multi} chooses $\boldsymbol{S}(i, A, \mathbb{H}):= A\bm{e}_i$ and $C=1$ (all units are in one cluster), which denotes the global proportion of treatment in each time $t$. 
    % They only consider the fixed best arm $S^* \in \{0,1\} \subseteq \{0,1/N,2/N,...1\}$. 
    % \footnote{Noteworthy, another difference is that~\citet{jia2024multi} considered the design-based setting where $\eta_{i,t}= 0$ and the potential outcomes in each time $t$ are different. It is beyond the scope of the current discussion, and we leave it for future work.}. 
    Additionally, the exposure mapping and clustering technique could also be traced back to the offline setting. \textbf{Example (iv)}
    Suppose $\forall j \in \U$, $\sum_j h_{ij} > 0$. We can choose $\boldsymbol{S}(i, A, \mathbb{H}) := \bone\{{\sum_{j\in\U} h_{ij} a_{j}}/{\sum_{j\in\U} h_{ij}} \in [0,\frac{1}{2})\}$ inherited from the literature of offline causality~\citep{leung2022causal, gao2023causal}. They require approximate neighborhood interference and their objective is to explore the influence of the treatment assignment proportion among all neighborhoods of each unit, which is still under-explored in the online learning scenario (we refer readers to experiments in Appendix~\ref{ex_app}).
% \end{example}\label{ex_2}
\textbf{Example (v)} For a supplement, we point out that the clustering strategy could also be traced back to the offline setting, which is also our special case:
 \citet{viviano2023causal, zhang2023individualized} considered the clustering-based setting $\boldsymbol{S}(i, A, \mathbb{H}):= A\bm{e}_i$, in which only considers the exposure arm set $\{0,1\}^{\mathcal{C}}$. Specifically, \citet{viviano2023causal} focuses on the cluster-wise Bernoulli design while \citet{zhang2023individualized} further assumes that the interference only occurs within clusters instead of across clusters.

In these examples, they all satisfy $\mathcal{U}_{\mathcal{E}} =\mathcal{U}_{\mathcal{C}} \cap \mathcal{U}_{\mathcal{O}} \neq \emptyset$. We provide more justification for it in the next section and Appendix~\ref{optimize_perspective}.
% After preparation, we establish the trade-off formulation. 

\subsection{Goal: estimation-regret trade-off} We introduce the goal of the trade-off between the regret efficiency and statistical power of reward gap estimation. Average treatment effect (ATE) between exposure super arm $S_i$ and $S_j$ is defined as the reward gap \citep{simchi2023multi}: 
% \begin{equation}
% \begin{aligned}
$\Delta^{(i,j)} := \frac{1}{N}\sum_{i^\prime \in \mathcal{U}} \big(\Tilde{Y}_{i^\prime}(S_i ) - \Tilde{Y}_{i^\prime}(S_j)\big), \text{where~}S_i, S_j \in \U_\bbE.$
% \end{aligned}\label{}
%     \end{equation}
It is a generalized definition compared with the most relevant literature~\citep{jia2024multi,agarwal2024multi, simchi2023multi} when considering ATE ({specifying the exposure mapping function as in Table ~\ref{compare} of Appendix~\ref{app_review}). % Recall that we denote the optimal arm as $S^* = \argmax_{S\in \U_\bbE } \frac{1}{N}\sum_{i \in \U } \Tilde{Y}_i(S  )$ as above. On this basis, we briefly denote $\Delta^{i}$ as the ATE (expected reward gap) between the optimal arm and exposure super arm $S_i$, namely, $\frac{1}{N}\sum_{i^\prime \in\U} (\Tilde{Y}_{i^\prime}(S^*) - \Tilde{Y}_{i^\prime}( S_i))$, following~\citet{simchi2023multi}.
% Moreover, generalized from~\citet{simchi2023multi}, the policy for decision makers is from the history $\mathcal{H}_{t-1}$. The difference is that to get the $N$-dimensional reward in time $t' \in [t-1]$ from the action $A_{t-1}$, people use exposure mapping as a medium. Specifically, designers select specific arm $A_t$ to receive the exposure arm $\{\textbf{S}(i, A_t, \mathbb{H})\}_{i \in \U}$, and then compute the reward in~\eqref{exposure}. 
We use $\hat{\Delta}^{(i,j)} := \{\hat{\Delta}^{(i,j)}_t\}_{t \geq 1}, \hat{\Delta} := \{\hat{\Delta}^{(i,j)}\}_{S_i,S_j \in \mathcal{U}_{\mathcal{E}}}$ to identify a sequence of adaptive admissible estimates of $\Delta^{(i,j)}$. The total design of an MAB experiment could be represented by the vector $\{\pi, \hat{\Delta} \}$. Our final goal is to portray the mini-max trade-off:

\begin{equation}
\begin{aligned}
\min_{\{\pi, \hat{\Delta}\}} \max_{\nu \in \mathcal{E}_0} \big( \bR_{\nu}(T, \pi),  e_\nu(T, \hat{\Delta}^{})  \big), \text{~where~}e_\nu(T, \hat{\Delta}^{}) := \max_{S_i,S_j \in \U_\bbE} \mathbb{E}\big[\big|\Delta^{(i,j)} - \hat\Delta_T^{(i,j)}\big|\big].
\end{aligned}\label{tradeoff}
\end{equation}
Given any feasible $\nu$, $\bR_{\nu}(T, \pi)$ is associated with $\pi$, while $ e_\nu(T, \hat{\Delta}^{})$ is associated with $\hat{\Delta}$. Due to the complicated relation between $\pi$ and $\hat{\Delta}$ w.r.t. the history $\mathcal{H}_t,$ $t \in [T]$, especially in the network interference setting, this multi-objective optimization is quite challenging. 
% \begin{remark}
%     {Other metric can also be selected such as the confident interval to quantify the performance of estimator and re-establish the trad-off. We defer such extension to the future work.}
% \end{remark}
% Here we adopt $\mathcal{R}_{\nu}$ instead of $\mathcal{R}$ to emphasize the regret computation is based on specific MAB instance $\nu$. 
For preparation, we define what is the ``{best}'' pair of $\{\pi, \hat{\Delta}\}$ via the following definition of \textit{front}:
\begin{definition}[Front and Pareto-dominate]\label{def_pareto}
    For a given pair of $\{\pi, \hat{\Delta}\}$, we call a set of pairs $(\bR,e)$ as a \textit{front} of $\{\pi, \hat{\Delta}\}$, denoted by $\mathcal{F}(\pi,\hat{\Delta})$, if and only if (i)
    [Feasible instances exists]  $ \mathcal{V}_0 :=\Big\{  \nu_0 \in \mathcal{E}_0: \Big(\sqrt{\bR_{\nu_0}(T, \pi)} , e_{\nu_0}(T, \hat{\Delta}^{})\Big) = (\bR,e)\Big\} \neq \emptyset$, and (ii)[instances in $\mathcal{V}_0$ are the best]  $ {\nexists}\nu \in \mathcal{E}/ \mathcal{V}_0, s.t. \exists \otimes \in \{K,T\}, (R, e) \preccurlyeq_{\otimes}    \big(\sqrt{\bR_{\nu}(T, \pi)}, e_{\nu}(T, \hat{\Delta}^{})\big)$. 
    % \begin{itemize}
    %     \item[(i)][Feasible instances exists]  $ \mathcal{V}_0 :=\Big\{  \nu_0 \in \mathcal{E}_0: \Big(\sqrt{\bR_{\nu_0}(T, \pi)} , e_{\nu_0}(T, \hat{\Delta}^{})\Big) = (\bR,e)\Big\} \neq \emptyset$.
    %     \item[(ii)][instances in $\mathcal{V}_0$ is the best]  $ {\nexists}\nu \in \mathcal{E}/ \mathcal{V}_0, s.t. \exists \otimes \in \{K,T\}, (R, e) \preccurlyeq_{\otimes}    \big(\sqrt{\bR_{\nu}(T, \pi)}, e_{\nu}(T, \hat{\Delta}^{})\big)$. 
    % \end{itemize}
   We claim $\{\pi, \hat{\Delta}\}$ Pareto-dominate another solution $\{\pi', \hat{\Delta}'\}$ if $\forall (\mathcal{R}, e) \in \mathcal{F}(\pi, \hat{\Delta})$, $\exists (\mathcal{R}', e') \in \mathcal{F}(\pi', \hat{\Delta}')$, such that $\forall \otimes \in \{K,T\}$, either (i) $\mathcal{R} \preccurlyeq_\otimes \mathcal{R}', e \prec_\otimes e'$ or (ii) $\mathcal{R} \prec_\otimes \mathcal{R}', e \preccurlyeq_\otimes e'$\footnote{Intuitively speaking, if we denote the region formed by $\mathcal{F}(\pi, \hat{\Delta}), \mathcal{F}(\pi', \hat{\Delta}')$, X-axis and Y-axis in the first quadrant as \texttt{Region}$(\pi, \hat\Delta)$, \texttt{Region}$(\pi', \hat\Delta')$, respectively. Then $\{\pi, \hat{\Delta}\}$ Pareto-dominate $\{\pi', \hat{\Delta}'\}$ means \texttt{Region}$(\pi, \hat\Delta) \subseteq \texttt{Region}(\pi', \hat\Delta')$.}.
\end{definition}
We formalize the definition of \textit{front} in the symbol of order $ \preccurlyeq_{\otimes}, \prec_\otimes$. {e.g.,  $(a,b) \preccurlyeq_{\otimes} (c,d), e \prec_\otimes f, g\preccurlyeq_\otimes h$ denotes $(a \leq c, b \leq d), e < f, g\leq h$ when we only consider the parameter with respect to $\otimes \in \{K,T\}$ sufficiently large and omit any other parameter.}
Finally, Pareto-optimality is identified according to the Pareto-dominance in Definition~\ref{def_pareto} as follows.
\begin{definition}[{Pareto-optimal and Pareto Frontier}]\label{frontier}
    A feasible pair $(\pi^*, \hat{\Delta}^*)$ is claimed to be Pareto-optimal when it is not Pareto-dominated by any other feasible solution. Pareto Frontier $\mathcal{P}$ is denoted as the envelope of fronts of all Pareto-optimal solutions. 
\end{definition}
For example, according to Definition~\ref{frontier}, $\{\pi_i, \hat{\Delta}_i\}_{i\in [3]}$ is not dominated by each other in Figure~\ref{fig_optimal}. For more intuitive comprehension for practitioners, we provide the closed-form mathematical formulation in the following section.
\begin{figure}[t]
\centering  %图片全局居中
\subfigure[Our general result under interference.]{
\label{fig1}
\includegraphics[scale = 0.21]{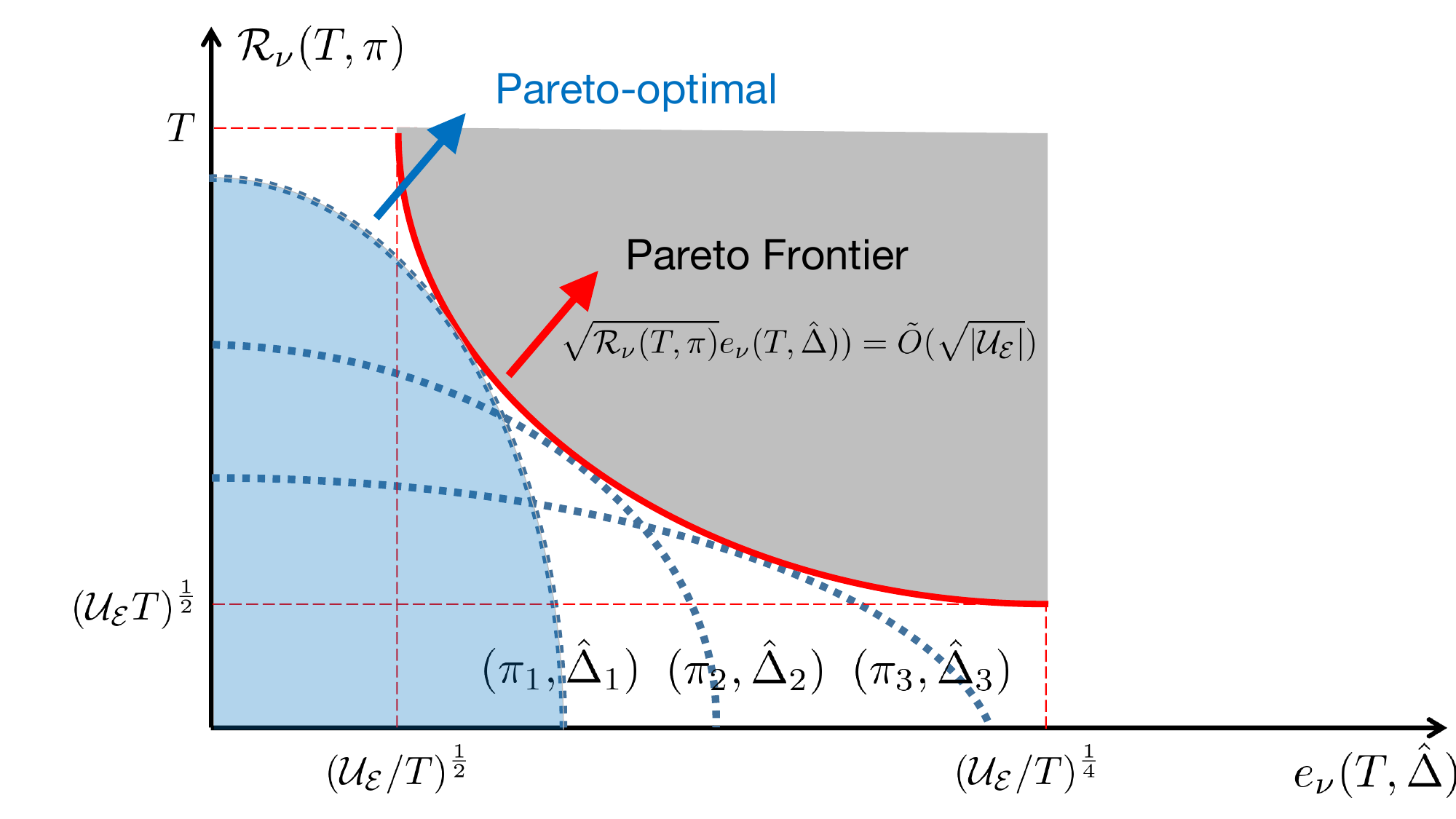}
}
\subfigure[The comparison with the baseline without interference.]{
\label{fig2}
\includegraphics[scale = 0.21]{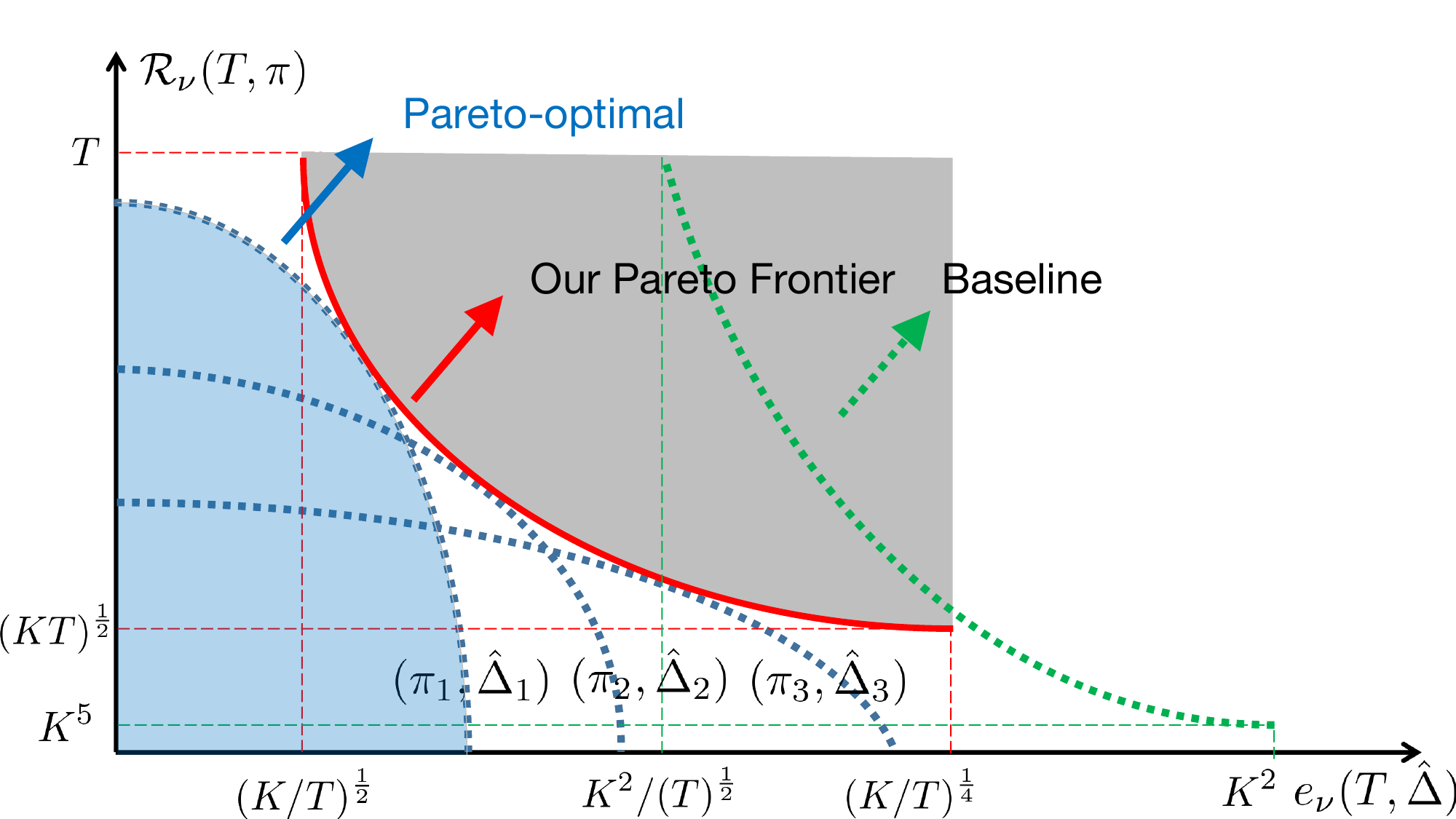}}
\caption{Pareto-optimality. (a) We use three blue fronts (first quadrant) to show three different MAB algorithms $\{\pi_i, \hat{\Delta}_i\}_{i\in [3]}$, e.g., the blue regions represent the regrets and estimation errors that can be realistically achieved in all kinds of instances given $\{\pi_1, \hat{\Delta}_1\}$. MAB algorithm is Pareto-optimal if and only if its blue front is tangent to the Pareto Frontier (red) (otherwise, it is intersecting with the grey region). (b) The green line represents the baseline in~\citet{simchi2023multi}, which loses the Pareo-optimality concerning arm space.}
\label{fig_optimal}
\end{figure}
\section{Pareto-optimality}\label{pareto}
In the above section, we introduce the motivation and establishment of our \texttt{MAB-N} and then construct the mini-max trade-off problem along with the Pareto-optimality property. In this section, we explore in detail the lower bound of such trade-off and the geometric structure of Pareto optimality. According to the Definition~\ref{def_pareto}-\ref{frontier}, in the following text, our analysis upon optimality mainly focuses on the individual arm space $K$ and the time horizon $T$. Here $K$ is included in the exposure arm space $\mathcal{U}_{\mathcal{E}}$. Other parameters, such as $N$, are seen as a pre-fixed constant. We first introduce the following condition to restrict the fairly broad relationship between parameters.

\begin{condition}\label{condition_in}
Exposure mapping $\bS$ and clusters $\mathcal{C}$ should satisfy $2 \le \vert \U_\mathcal{E}\vert \le T $. 
\end{condition}
Condition~\ref{condition_in} restricts to the case where $T$ is relatively large with pre-specified non-empty $\mathcal{U}_{\mathcal{E}}$, which is inherently verifiable, adjustable and relevant. Regardless of any pre-fixed $\mathbb{H}$, we could manually design legitimate~\eqref{exposure_def} and clusters to fit Condition~\ref{condition_in}. It is the weakest condition to date, without additional restriction upon network topology, compared to the previous literature mentioned in the above section. Additional justification on exposure mapping and feasibility of model conditions are in Appendix~\ref{app_discussion} and Appendix~\ref{optimize_perspective}. Under such conditions, we establish a general lower bound when simultaneously considering the regret and estimation error. 

% In the following paper, we start from the case $K=2$, i.e., $i,j \in \{0,1\}$. 

\begin{theorem}\label{trade-off}
   Given any $\mathbf{S}$ and $\mathcal{C}$ that satisfies Condition~\ref{condition_in}.
   Given any online decision-making policy $\pi$, the trade-off between the regret and the estimation exhibits
    \begin{equation}
    \begin{aligned}
    \inf _{\hat{\Delta}_T} \max _{\nu \in \mathcal{E}_0} \Big( \sqrt{\bR_{\nu}(T, \pi)}  e_\nu(T, \hat{\Delta}^{}) \Big) =  \Omega_{K,T}\Big(\sqrt{|\mathcal{U}_{\mathcal{E}}|}\Big).
    \end{aligned}
    \end{equation}
\end{theorem}
We use the subscript $\{K, T\}$ to emphasize that the order just corresponds to these two parameters and omit the subscript in the following text.

\noindent \textbf{The challenge of the proof} 
The core idea involves constructing two carefully designed multi-armed bandit instances, \(\nu_1\) and \(\nu_2\), such that any estimator \(\hat{\Delta}_T\) faces challenges in simultaneously achieving low regret and high estimation accuracy across both instances. This difficulty is divided into three parts: (i) Regarding the goal, unlike the regret lower bound analysis in classic multi-armed bandit problems~\citep{Lattimore2020BanditA}, we employ statistical hypothesis testing to bridge these two goals, rather than analyzing worst-case regret in isolation. (ii) Concerning instance construction, compared to \citet{simchi2023multi}, constructing two distinct instances is challenging due to the interference affecting the entire system, making it difficult for an algorithm’s regret or estimation behavior to differ significantly. (iii) From an information-theoretic perspective, the correlated structure complicates the issue. The networked nature of exposure rewards necessitates a refined divergence measure that accounts for shifts in probability mass across dependent actions, such as when applying the Kullback-Leibler inequality. 

\noindent \textbf{The sketch of the proof.} We defer the detailed proof in Appendix~\ref{app_tradeoff}. To tackle these challenges, we carefully construct a pair of instances $\{\nu_1, \nu_2\}$ via slighting perturbing the reward of $Y(A_t)$ compatible with specific exposure arms: we let $\nu_1$ as $Y_i(A):= f_i(A) \in (\varepsilon_0,1-\varepsilon_0), \varepsilon_0 \in (0,1), r_{i,t}(A) \in \{-1,1\}$. It means
         $r_{i,t}(A) = \text{Rad}(\frac{1-f_i(A)}{2}, \frac{1+f_i(A)}{2})$.
Moreover, We establish $\nu_2$ as:
        \begin{equation}
        \begin{aligned}
          r'_{i,t}(A) := \begin{cases}
                r_{i,t}(A) &\forall A\text{~satisfying~} \mathbb{P}_{}(A_t = A \mid S ) = 0.   \\
               \text{Rad}(\frac{1-f_i(A)+\alpha}{2}, \frac{1+f_i(A)-\alpha}{2}) & \forall A\text{~satisfying~} \mathbb{P}_{}(A_t = A \mid S )  > 0.
          \end{cases}
        \end{aligned}
        \end{equation}
with $\alpha>0$ sufficiently small, and $S$ is specifically selected. Conducting the information-theoretic argument, we prove
\[
\inf _{\hat{\Delta}_T} \max _{\nu \in \mathcal{E}_0} \mathbb{P}_\nu\left(\max_{i,j \in \U_{\mathcal{E}}}|\hat{\Delta}^{(i,j)}_T-\Delta_\nu^{(i,j)}| \geq \frac{\alpha}{2} \right)  \geq \frac{1}{2}\bigg[1-\sqrt{\frac{1}{2}  q' N \alpha^2 \frac{\bR_{\nu_1}(T, \pi)}{|\mathcal{U}_{\mathcal{E}}| }}\bigg].
\]
Here $q'$ is a constant. Such inequality bridges the relationship between the statistical power and regret efficiency under these two instances and thus induces the final lower bound in Theorem~\ref{trade-off}.

Theorem~\ref{trade-off} states that for any given policy $\pi$, there always exists at least one hard MAB instance $\nu$, in which no matter what legitimate $\mathbf{S}$, $\bC$, and estimator $\hat{\Delta}_T$ we choose, the lower bound $\Omega(\sqrt{|\mathcal{U}_{\mathcal{E}}|})$ always holds. In other words, there are always challenging instance $\nu$ such that $e_\nu(T, \hat{\Delta}) = \Omega_{K,T}({\sqrt{|\mathcal{U}_{\mathcal{E}}|}}/{\sqrt{\bR_{\nu}(T, \pi) }})$. We take examples considering \textit{the worst case of} $\nu$: according to the fact $\mathcal{R}_{\nu}(T,\pi) = O(T)$, Theorem~\ref{trade-off} states that the worst estimation error is at least $\Omega(({|\mathcal{U}_{\mathcal{E}}|}/{{T}})^{\frac{1}{2}})$ and could not be further decreased; stepping forwards, as we will show in~\Cref{alg} that our proposed \texttt{MAB-N} algorithm's regret is upper bounded by $O(\sqrt{|\mathcal{U}_{\mathcal{E}}|T})$, then Theorem~\ref{trade-off} additionally states that the worst estimation error of our algorithm will be ideally at least $(|\mathcal{U}_{\mathcal{E}}|/T)^{\frac{1}{4}}$ without need of further implementation. In sum, Theorem~\ref{trade-off} serves as a \textit{free lunch}, enabling practitioners to perform interactive inference and prediction regarding the trade-off between the algorithm's regret efficiency and statistical power. A natural question is what is the relationship between the lower bound and the Pareto-optimality? We provide the following closed-form for Pareto Frontier following the lower bound in Theorem~\ref{trade-off}.

\begin{theorem}\label{thm_pareto}
     Following the condition in Theorem~\ref{trade-off}, a feasible pair $\{\pi, \hat{\Delta}\}$ is Pareto-optimal \textit{if the pair satisfies~} $\max _{\nu \in \mathcal{E}_0} \big( \sqrt{\bR_{\nu}(T, \pi)} e_\nu(T, \hat{\Delta}^{}) \big) = \Tilde{O}\big(\sqrt{|\mathcal{U}_{\mathcal{E}}|}\big).$ The Pareto Frontier is represented as $\mathcal{P} = \big\{ (\bR_{\nu}(T, \pi),e_\nu(T, \hat{\Delta}^{}) ):  \sqrt{\bR_{\nu}(T, \pi)} e_\nu(T, \hat{\Delta}^{})  = \Tilde{O}(\sqrt{|\mathcal{U}_{\mathcal{E}}|})  \big\}$.
\end{theorem}
Theorem~\ref{thm_pareto} establishes the sufficiency condition for the Pareto-optimal property. We also analyze the necessity conditions in Appendix~\ref{app_optimal}. For a visual representation, readers are referred to Figure~\ref{fig_optimal}, which illustrates the Pareto-optimal pairs \({\pi,\hat{\Delta}}\) (blue region) and the Pareto Frontier (red line). Theorems~\ref{trade-off} and \ref{thm_pareto} are applicable to any complex network topology \(\mathbb{H}\) under mild conditions on exposure mapping (Condition~\ref{condition_in}). These results not only generalize non-trivial trade-offs under network interference but also enhance the degenerated results without interference. Specifically, when compared to the setting of \citet{simchi2023multi}, (i) we advance the Pareto-optimality trade-off concerning arm space, and (ii) we eliminate their additional assumption on ATE, specifically that \(\hat{\Delta}^{i,j} = \Theta(1)\). Furthermore, our reward \(r_t\) is not constrained to the interval \([-1,1]\), allowing for unbounded values.

\section{Algorithm}\label{alg}

To achieve the Pareto-optimality trade-off outlined in Section~\ref{pareto}, we introduce the advanced \textbf{U}pper \textbf{C}onfidence \textbf{B}ound algorithm with \textbf{T}wo \textbf{S}tages under \textbf{N}etwork interference (\texttt{UCB-TSN}). The algorithm aims to bound both ATE and regret simultaneously. Our \texttt{UCB-TSN} operates in two phases: (i) uniformly exploring the super exposure arm space to generate an estimated ATE, and (ii) applying the UCB exploration strategy to minimize regret. Initially, we demonstrate that phase (i) effectively reduces the estimation error, as detailed below.

\begin{theorem}[ATE estimation upper bound] \label{stochasticATE} Following the condition in Theorem~\ref{trade-off}. If $T_1 \ge \vert \U_\bbE \vert$, for any $S_i \not = S_j \in \U_\mathcal{E}$, the ATE estimation error of \text{\texttt{UCB-TSN}} can be upper bounded as $  \bE\big[\vert \hat{\Delta}_T^{(i,j)} - \Delta^{(i,j)} \vert\big] = \tilde{O}\big(\sqrt{{\vert \U_\bbE \vert}/{T_1}}\big)$.   
\end{theorem}
Theorem~\ref{stochasticATE} asserts that uniform exploration in phase (i) aids in estimating the ATE. This is intuitive, as \texttt{UCB-TSN} explores the exposure action space using a round-robin approach. Provided that the practitioner selects \(T_1 = \Omega(T^{\alpha})\) for \(\alpha \in (0,1)\), the ATE estimation is consistent. Following the uniform exploration in phase (i), phase (ii) focuses on identifying the optimal arm, leading to the convergence of the overall regret.

% \begin{itemize}
    % \item $N$ is added to the de-nominator when one-to-one mapping;
    % for instance, generalized su, $N\rightarrow +\infty$, $\mathcal{U}_{\mathcal{E}} = d_s^C \ll N$, the error converge to 0 (another intuitive explanation noise degenerate to 0, the online degenerate to the offline;); for instance, Agarwal arm space diverge..$K^N/N$ impractical result; further validates Proposition~\ref{negative} (not only regret is impractical; but also estimation error without assumption.) Consider the exposure map as a subset selection instead of the information wrapping.
% \end{itemize}

% \zhiheng{discussion on NN:  2 when N→+∞N \rightarrow +\infty,  3 when we choose the cluster-switchback policy on the original/exposure super arm; is the one-one mapping; }

% \zhiheng{discussion }

\begin{theorem}[Regret upper bound] \label{stochasticregret} Following the condition in Theorem~\ref{trade-off}. With $\delta = {1}/{T^2}$ and $T_1 \ge \vert \U_\bbE \vert$, the regret of \texttt{UCB-TSN} can be upper bounded as $\bR(T,\pi) = \tilde{O}\big(\sqrt{\vert \U_\bbE \vert T} + T_1\big)$.
\end{theorem}
Theorem~\ref{stochasticregret} claims the regret could converge as $o(T)$, accommodating with well-selected $T_1$, such as $T_1 = \sqrt{|\mathcal{U}_{\mathcal{E}}|T}$. Theorem~\ref{stochasticregret} is consistent with Proposition~\ref{negative} when we omit phase (i), i.e., $T_1 = 0$ and reserve phase (ii). By the combination of Theorem~\ref{stochasticATE}-\ref{stochasticregret}, we claim the Pareto-optimality as stated in~\Cref{pareto} in our \texttt{UCB-TSN} as follows.
% \zichen{If we set T1=0T_1 = 0, this result can match to the hard case when |\U\bbE|=KN\vert \U_\bbE \vert = K^N. }
% \zhiheng{1 match ucb; 2 outperform chonghuan}
\begin{corollary}[Trade-off result]\label{trade-off1}
 Following the condition in Theorem~\ref{trade-off}. Set $T_1 \ge \sqrt{\vert \U_\bbE \vert T}$, for all $\nu\in\bbE_0$, \texttt{UCB-TSN} can guarantee $e_\nu(T,\hat{\Delta})\sqrt{\bR_\nu(T,\pi)} =  \tilde{O}(\sqrt{\vert \U_\bbE \vert})$.
\end{corollary}
Corollary~\ref{trade-off1} states that under a stricter but still mild condition upon the uniform exploration process $T_1$ (since $\sqrt{|\U_\bbE|T} \ge |\mathcal{U}_{\mathcal{E}}|$ under Condition~\ref{condition_in}), \texttt{UCB-TSN} could achieve the Pareto-optimal property according to Theorem~\ref{trade-off}. Simulation results are provided in the Appendix \ref{ex_app} to validate its effectiveness. Moreover, we comment on the order of $N$ in Appendix~\ref{app_exposure_sec}.

\paragraph{Comparison with the baseline algorithm.}
\label{comparison} To facilitate the fair comparison, we consider the degenerated case as in~\citet{simchi2023multi}, where we choose $N=1, \vert \mathcal{U}_{\mathcal{E}}\vert = K \geq 2$ in our \texttt{UCB-TSN}. Here $\mathcal{K}$ corresponds to $\U_\bbE$.

We compare the regret in (i) and estimation in (ii). (i) For the regret, they proposed their \texttt{EXP3EG} which guarantees the regret upper bound as $\mathcal{R}_{{\nu}}(T,\pi) = \tilde{O}(K^5 + T^{1-\alpha})$, where $\alpha \in [0,1]$\footnote{In their paper, $\mathcal{R}_{{\nu}}(T,\pi) = {O}\big(\sum_{A \in \mathcal{K}/\{A^*\}} {K^4 log(T)} +  T^{1-\alpha}log(T) \big) = \tilde{O}(K^5 + T^{1-\alpha}).$ Here $A^*$ denotes the best super arm.}. Such result is build upon their assumption $\frac{1}{N}\sum_{i^\prime \in \mathcal{U}} \big(\Tilde{Y}_{i^\prime}(S^* ) - \Tilde{Y}_{i^\prime}(S_i)\big) =\Theta(1)$ for all $S_i\not=S^*$. In this single-agent setting with such assumption, it should be pointed out that our regret upper bound in Theorem~\ref{stochasticregret} could be naturally strengthened to ${\tilde{O}}\big(K + T_1\big) $ (refer to our instance dependent regret upper bound in Lemma \ref{dependent_regret_stochastic} in the Appendix). Thus our regret upper bound is strictly stronger than~theirs if we force $T_1 = O(T^{1-\alpha})$. (ii) For the estimation error, they state that ATE could be upper bounded by~{$e_{\nu}(T,\pi) = \Tilde{O}({K^2}{T^{-\frac{1-\alpha}{2}}} )$.} Therefore our estimation error in Theorem~\ref{stochasticATE}, i.e., $\tilde{O}(\sqrt{{|\mathcal{U}_{\mathcal{E}}|}/{T_1}}) = \tilde{O}(\sqrt{{K}/{T_1}})$ is strictly stronger than theirs since it is legitimate to force $T_1 = T^{1-\alpha} \vee \vert \U_\bbE \vert$. Such strict improvement (i)-(ii) is illustrated in Figure~\ref{fig_optimal}. It validates the statements under Theorem~\ref{thm_pareto} that we achieve the Pareto optimality with respect to time period $T$ and additionally, the exposure super arm space $|\mathcal{U}_{\mathcal{E}}|$. 

% Moreover,~\citet{simchi2023multi} only focuses on the bounded noisy reward within $[-1,1]$ while we extend the setting to the more general 1 sub-Gaussian
% noise for each arm.
% \zhiheng{cont'd}

% \zichen{In chonghuan's setting, where |\U\bbE|=K\vert \U_\bbE \vert = K, N=1N = 1, and T>>KT >> K. The UCB-TS is significantly better than chonghuan's algorithm}

% \section{Theoretical analysis}

% \subsection{Regret analysis of algorithm}
\begin{algorithm*}[t]
\renewcommand{\algorithmicrequire}{\textbf{Input:}}
\renewcommand{\algorithmicensure}{\textbf{Output:}}
	\caption{UCB-Two Stage-Network (\texttt{UCB-TSN})}\label{alg1}
	\begin{algorithmic}
         \STATE \textbf{Input:} arm set $\A$, time $\{T_1,T\}$, unit number $N$, exposure super arm set $\U_\mathcal{E}$, estimator set $\{\hat{R}_0(S) = 0\}_{S \in \U_\bbE}$, $\{\N_0^S = 0\}_{S \in \U_\bbE}$, $\{\text{UCB}_{0,S} = 0\}_{S \in \U_\bbE}$, counter $k = 1$
        \FOR{$t = 1 : T_1$}
        \STATE Select exposure super arm $S_t = S_k$ and implement $\texttt{Sampling}(S_t)$
        \STATE Set $k = k + 1$ if $k + 1 \le \vert \U_\bbE \vert$, else set $k = 1$
        \ENDFOR
         \STATE For all $S_i,S_j \in \U_\mathcal{E}$, $S_i \not = S_j$, output $\hat{\Delta}_T^{(i,j)} = \hat{R}_{T_1}(S_i) - \hat{R}_{T_1}(S_j) $
        \FOR{$t = T_1 + 1 : T$}
         \STATE Select $S_t = \arg\max_{S\in\U_\bbE} \text{UCB}_{t-1,S}$ and implement $ \texttt{Sampling}(S_t)$
        \ENDFOR\\
        \# Parameter~1: $\N_S^{t} =  \sum_{t^\prime = 1}^{t} \bone\{ S_{{t}^\prime} = S \}$\\
        \# Parameter~2: $\hat{R}_t(S) =  \big({\hat{R}_{t-1}(S) \N_S^{t-1} + \bone\{S_t = S\} \frac {1}{N}\sum_{i \in \U}\tilde{r}_{i,t}(S)}\big)/{\N_{S}^t}$ \\
        \# Parameter~3: $\text{UCB}_{t,S} = \hat{R}_{t}(S) + \sqrt{{18\log(1/\delta)}/{{\N_S^{t}}}         }$
	\end{algorithmic}  
\end{algorithm*}

\begin{algorithm*}[t]
\renewcommand{\algorithmicrequire}{\textbf{Input:}}
\renewcommand{\algorithmicensure}{\textbf{Output:}}
	\caption{\texttt{Sampling}}
	\begin{algorithmic}
        \STATE \textbf{Input:} $S_t$
                \STATE Derive the set $\{Z_{l'}\}_{l'\in[l]}$ such that $\{\bS(i,Z_{l'},\bbH)\}_{i\in\U} = S_t$, $\forall l'\in[l]$; sample $A_t$ from set $\{Z_{l'}\}_{l\in[l']}$ based on $\mathbb{P}(A_t = A \mid S_t)$, pull $A_t$,
        and observe reward $\{\tilde{r}_{i,t}(S_t) = r_{i,t}(A_t)\}_{i \in \U}$
        \end{algorithmic}
\end{algorithm*}

\section{Extension to adversarial setting}\label{adversarial_sec}

\paragraph{The adversarial setting.} {{We cover~\citet{simchi2023multi}'s adversarial setting when considering trade-offs}}. We consider $r_{i,t}(A_t) = Y_i(A_t) + f_t + \eta_{i,t}$, where $\eta_{i,t}$ is i.i.d. zero means noise. In addition to the standard setting in the preliminaries, there is an $f_t$, a pre-specified function w.r.t. period $t$, which is an adversarial noise. We suppose $r_{i,t}(A) \in [0,1]$ for all $i\in\U$, $A\in\bK^\U$ and $t\in[T]$. It is also easy to verify that $\tilde{r}_{i,t}(S) \in [0,1]$ for all $t\in[T]$, $S_i\in\U_\bbE$, $i\in\U$ and $\bE[\tilde{r}_{i,t}(S)] = \tilde{Y}_i(S) + f_t$.  Motivated by the fact that the UCB algorithm discussed in the previous section cannot be applied directly in this context, we provide the advanced \texttt{EXP3-TSN} algorithm for substitution. The pseudo-code and details of the \texttt{EXP3-TSN} are provided in the Appendix. We provide the estimation error, regret, and trade-off in Theorem~\ref{thm_adversarial}.   

\begin{theorem}[Pareto-optimality trade-off in the adversarial setting]\label{thm_adversarial}  Following the condition in Theorem~\ref{trade-off}, let $\mathcal{T}(t) \equiv {(2\vert \U_\bbE \vert + 1)^2\log(t \vert \U_\bbE \vert^2)}/{2(e-2)\vert \U_\bbE \vert}$, then

(i)\texttt{[ATE~estimation]} 
    % Given any $\mathbf{S}$ and $\mathcal{C}$ that satisfy Condition~\ref{condition_in}. 
    Suppose $T \ge \mathcal{T}(T)$ and $T_1 \ge \mathcal{T}(T_1)$. For any $S_i \not = S_j$, the ATE estimation error of the \texttt{EXP3-TSN} can be upper bounded as in Theorem~\ref{stochasticATE}. 
    
    (ii)\texttt{[Regret]} Stepping back, if we only suppose $T \ge \mathcal{T}(T)$, then the regret of \texttt{EXP3-TSN} could be upper bounded as in Theorem~\ref{stochasticregret}. 
    
    (iii)\texttt{[Pareto-optimality]} Stepping forward, additionally set $T_1 \ge \mathcal{T}(T_1) \vee \sqrt{\vert 
\U_\bbE \vert T}$. then \texttt{EXP3-TSN} can also guarantee the Pareto-optimality trade-off, i.e., $e_\nu(T,\hat{\Delta})\sqrt{\bR(T,\pi)} = \tilde{O}(\sqrt{\vert \U_\bbE \vert})$.
\end{theorem}
% \begin{theorem}\label{theorem2}
%   Given any instance that satisfy $T \ge \mathcal{T}(T)$ and $\vert \U_\bbE \vert \ge 2$. Set $T_1 \ge \mathcal{T}(T_1) \vee \sqrt{\vert 
% \U_\bbE \vert T}$. \texttt{EXP3-TS} can guarantee $e_\nu(T,\hat{\Delta})\sqrt{\bR(T,\pi)} = \tilde{O}(\sqrt{\vert \U_\bbE \vert})$.
% \end{theorem}
% \paragraph{Justification of these conditions}
Theorem~\ref{thm_adversarial} states that under additional mild conditions, i.e., $T \geq \mathcal{T}(T)$ and $T_1 \geq \mathcal{T}(T_1)$\footnote{Since $\mathcal{T}(t) = O({\vert \mathcal{U}_{\mathcal{E}}\vert}log(|\U_\mathcal{E}|t))$, such conditions are natural to satisfy given that $T$ is sufficiently large.}, the regret, ATE estimation error and the Pareto-Optimality trade-off could still keep their original form in Theorem~\ref{stochasticATE}-\ref{stochasticregret}. In such an adversarial setting, our result can also outperforms~\citet{simchi2023multi} with the same argument as in~\Cref{comparison}, and the discussion concerning the order of the node number $N$ aligns analogously.

% \begin{theorem}[Bounding the ATE estimation with high probability] \label{lemmaATEadv} For any $S_i \not = S_j$ and $S_i,S_j \in \U_\bO \cap \U_\bC$, with probability at least $1-\delta$, the ATE estimation error of the EXP3-TS can be upper bounded as follows:
% \begin{align}\label{ATEresults}
%     \vert \hat{\Delta}^{(i,j)} - \Delta^{(i,j)} \vert = O\Bigg(\sqrt{\frac{\vert \U_\mathcal{E} \vert}{T_1}} \log\Big(\frac{\vert \U_\mathcal{E} \vert}{\delta}\Big) \Bigg).
% \end{align} 
% \end{theorem}

% \begin{theorem}[Instance-dependent regret] \label{lemmaregretadv} With probability at least $1-\delta$, the regret of the EXP3-TS can be upper bounded as follows
% \begin{align}
%     \bR(T,\pi) = O \Bigg(K \sum_{i = 1, S_i \not = S^*}^{\vert \U_\mathcal{E}\vert}\frac{\log(\frac{\vert \U_\bO \cap \U_\bC \vert}{\delta})}{\Delta^{*,i}} + T_1\Bigg).
% \end{align}    
% \end{theorem}

% \begin{theorem}[Instance-independent regret]\label{theorem3adv} With probability at least $1 - \delta$, the regret of EXP3-TS can be upper bounded by 
%     \begin{align}
%         \bR(T,\pi) = O \Bigg( \sqrt{\vert \U_\mathcal{E} \vert T }\log\Big(\frac{\vert \U_\mathcal{E} \vert}{\delta}\Big) + T_1 \Bigg).
%     \end{align}
% \end{theorem}
% \zhiheng{T1 $\sqrt{T}$ order...}

% \begin{theorem}\label{theorem2}
%    For any MAB instance $v$, EXP3-T can guarantee $e_v(T,\hat{\Delta})\sqrt{\bR(T,\pi)} = \tilde{O}(1)$.
% \end{theorem}

% \section{Experiments}

\section{Conclusion and future work}\label{discussion} 

This paper establishes a generalized bandit framework under network interference via exposure mapping, balancing learning efficiency and statistical power through a Pareto-optimal trade-off between regret and estimation error. We introduce \texttt{UCB-TSN}, a theoretically grounded algorithm achieving this balance, validated experimentally (Appendix~\ref{ex_app}) and extended to more complex interference structures.   

Beyond this contribution, our work connects statistical and learning communities by framing network interference as a fundamental mechanism in interactive decision-making. The flexibility of exposure-based modeling, the stability of our optimality guarantees, and the extensibility of our framework make it a foundation for broader advances in sequential decision-making under structured dependence. Key future directions include: (i) Reinforcement Learning: Extending to sequential decision-making in networked environments, where interference shapes both short- and long-term rewards~\citep{tran2023inferring}. (ii) Bandit Variants and Statistical Inference: Addressing fully adversarial bandits, design-based experiments, and continual anytime-valid inference~\citep{howard2021time}. Exploring neural bandits and graph neural network-based models. (iii) Generalized Network Structures: Advancing toward dynamic, unknown, and heterogeneous network interference models, leveraging adaptive clustering techniques~\citep{zhao2024simple, viviano2023causal}.  These directions aim to lay the groundwork for efficient learning with statistical power under structured dependence. They will be continuously developed to be a bridge between these two communities for future research.

% \input{text/2-Preliminary_overview}

% \clearpage
\vskip 0.2in
\bibliography{sample}

\newpage
% \onecolumn
\appendix

% \part{Appendix} % Start the appendix part
% %	\begin{center}
%     %		\LARGE \bm {Appendix of XXX}
%     %	\end{center}
% \vspace{-20pt}
% \etocdepthtag.toc{mtappendix}
% \etocsettagdepth{mtchapter}{none}
% \etocsettagdepth{mtappendix}{subsection}
% \renewcommand{\contentsname}{}
% \tableofcontents

% \newpage

\begin{center}
   \Large{Supplementary Material for ``Online Experimental Design With Estimation-Regret
Trade-off Under Network Interference''}
\end{center}

Appendix~\ref{notation} summarizes key symbols in the main text for reference.

Appendix~\ref{app_review} provides a detailed literature review for better comprehension of the background.

Appendix~\ref{app_discussion} and~\ref{optimize_perspective} provide the justification for exposure mapping and model conditions. 

Appendix~\ref{ex_app} illustrates the experiments.

Appendix~\ref{app_exposure_sec} further analyzes the structure of the exposure mapping and the re-scaled noise.

Appendix~\ref{app_propostion1} provides the proof the Proposition~\ref{negative}.

Appendix~\ref{app_tradeoff}-\ref{app_optimal} contain the proof of Theorem~\ref{trade-off} and Theorem~\ref{thm_pareto}, respectively.

Appendix~\ref{app_alg} presents the proofs of Theorem~\ref{stochasticATE}-\ref{trade-off1} in~\Cref{alg}.

Appendix~\ref{app_additional_alg} provides an algorithm for Non-stochastic Settings.

Appendix~\ref{app_adver} delivers the proof of Theorem~\ref{thm_adversarial}. Finally, Appendix~\ref{app_auxiliary} includes the auxiliary lemmas.

% \clearpage
  % \title{\bf Supplementary Material for ``Adjusting auxiliary variables under approximate neighborhood interference''}
  % \date{}
  % \maketitle

\section{Notations}\label{notation}

\begin{center}
\scalebox{0.85}{
\begin{tabular}{l|l}
\toprule[1.5pt]
    $\bK$ & Real arm set\\
    $K$ & Number of arms\\
    $\U$ & Unit set \\
    $N$ & Number of units \\
    $\bC$ & Cluster set\\
    $C$ & Number of clusters\\
    $\nu$ & Instance\\
    $\bbE_0$ & Set of the legitimate instance\\
    $\pi$ & Learning policy\\
    $\bR(T,\pi)$ & Cumulative regret of policy $\pi$\\
    $T$ & Time horizon\\
    $T_1$ & Length of the first exploration phase\\
    $Y_i(\cdot)$ & Potential outcome of unit $i$\\
    $\tilde{Y}_i(\cdot)$ & Exposure potential outcome of unit $i$\\
    $\bS(\cdot)$ & Exposure mapping\\
    $\bbH$ & Adjacency matrix\\
    $a_{i,t}$ & Action of unit $i$\\
    $s_{i,t}$ & Exposure action of unit $i$\\
    $A_t$ & Supper arm played $t$\\
    $S_t$ & 
    Exposure super arm played $t$\\
    $S^*$ & Optimal exposure super arm\\
    $d_s$ & Number of the exposure arm\\
    $\U_s$ & Exposure arm set\\
    $\U_\bC$ & Cluster-wise switchback exposure super arm set\\
    $\U_\bO$ & Set of exposure supper arm that can be triggered by real supper arm\\
    $\U_\mathcal{E}$ & Legitimate exposure super arm set \\
    $\tilde{r}_{i,t}(S)$ & Reward feedback of unit $i$ in round $t$ if exposure super arm $S$ is pulled\\
    $\Delta^{(i,j)}$ & ATE between $S_i$ and $S_j$\\
    $\Delta^i$ & ATE between $S^*$ and $S_i$
    \\
    $\hat{\Delta}_T^{(i,j)}$ & Estimated ATE between $S_i$ and $S_j$\\
    $\hat{R}_t(S)$ & Reward estimator of exposure super arm $S$\\
    $e_\nu(T, \hat{\Delta}^{})$ & Largest ATE estimation error\\
    $\N_{S}^t$ & Observation number of exposure super arm $S$ until round $t$\\
 \toprule[1.5pt]
    \end{tabular}}
\end{center}

% \newpage

\section{Literature Review}\label{app_review}

\begin{table}[t]
\centering%把表居中
\scalebox{0.67}{
\begin{tabular}{ccccc}%四个c代表该表一共四列，内容全部居中
 \toprule[1.5pt]
Interference-based MAB&Exposure mapping~($\boldsymbol{S}(i, A, \mathbb{H})$)&Action space ($|\mathcal{U}_{\mathcal{E}}|$)&Clusters ($\mathcal{C}$) & Estimation goal ($\Delta^{(i,j)}$) \\
\midrule%第二道横线 
\citet{simchi2023multi} &  $A$ & $K$ & $1$ & $Y(A_i) - Y(A_j)$\\
\citet{jia2024multi}&$A\bm{e}_i$ &$K$ & $1$ & $\frac{1}{N}\sum_{i' \in \mathcal{U}} (Y_{i'}(i*\bone_{N}) - Y_{i'}(j * \bone_{N}))$\\
\citet{agarwal2024multi}&$ A \bm{e}_{i} $&$K^{N}$ & $N$ & $\frac{1}{N}\sum_{i' \in \U} (Y_{i'}(A_{i}) - Y_{i'}(A_j))$ \\
{\texttt{MAB-N} (\textbf{Ours})}&General $\bS(i, A, \mathbb{H})$ & $O\big(|d_s|^C\big)$& $ C$ & $ \frac{1}{N}\sum_{i^\prime \in \mathcal{U}} \big(\Tilde{Y}_{i^\prime}(S_i ) - \Tilde{Y}_{i^\prime}(S_j)\big)$\\
 \toprule[1.5pt]
\end{tabular} }
\caption{{\texttt{MAB-N} surrogates the previous bandit under interference as special cases. Here $A_i, A_j \in \mathcal{K}^{\U}$, and $S_i, S_j \in \U_\bbE$.} We omit the subscript in~\citet{simchi2023multi} since it only considers sole individual.}%标题
\label{compare}
\end{table}

In this section, we present a literature review on network interference within the causality and bandit communities. Additionally, we discuss relevant variants of bandit problems. Finally, we provide a brief summary of recent advancements in the estimation-regret trade-off within the context of MAB.

\paragraph{Offline causality estimation under network interference.} In the current causality literature, interference is a well-known concept. It is a violation of the conventional ``SUTVA'' setting, representing that one individual's treatment would potentially affect another individual's outcome, which is relevant in practice. Current literature resort to clustering~\citet{zhang2023individualized,viviano2023causal} or exposure mapping~\citet{leung2022causal, leung2022rate, leung2023network}.

\paragraph{Bandit under network interference.} Previous attempts are being made to consider the multi-armed bandit problem upon network interference. \citet{agarwal2024multi} conduct the Fourier analysis to transform the traditional stochastic multi-armed bandit into a sparse linear bandit. However, in order to reduce the exponential action space, they made a strong assumption of sparsity for network structures, i.e., the number of neighbors of each node is manually upper limited. On the other hand, \citet{jia2024multi} analyzes the action space at the other extreme that considers an adversarial bandit setting and thus forces each node to a simultaneous equal arm. It does not consider that the optimal arm could differ for each node or subgroup. Moreover, \citet{xu2024linear} further considers the contextual setting under the specific linear structure between the potential outcome and the interference intensity.

\paragraph{Relevant bandit variants: multiple-play bandits, multi-agent bandits, combinatorial bandits, and multi-tasking bandits.} In bandit literature, the problem where a bandit algorithm plays multiple arms in each time period has been a subject of study for a long time. Our work is closely related to the \textit{multi-play bandit} problem, where the algorithm selects multiple arms in each round and observes their corresponding reward feedback \citep{anantharam1987asymptotically,uchiya2010algorithms,komiyama2015optimal,komiyama2017position,louedec2015multiple,lagree2016multiple,zhou2018budget,besson2018multi,jia2023short,wang2023pure}. Additionally, this is closely related to the \textit{multi-agent bandit} problem (including distributed and federated bandits), where multiple agents each pull an arm in every time period. By exchanging observation histories through communication, these agents can collaboratively accelerate the learning process. \citep{Hillel2013DistributedEI,Szrnyi2013GossipbasedDS,Wu2016ContextualBI,Wang2019DistributedBL,Li2022CommunicationEF,He2022ASA,wang2023pure}. Furthermore, our work is also connected to the \textit{combinatorial bandit} problem, where the action set consists of a subset of the vertices of a binary hypercube \citep{cesa2012combinatorial,chen2013combinatorial,chen2014combinatorial,combes2015combinatorial,qin2014contextual,kveton2015combinatorial,li2016contextual,saha2019combinatorial,wang2023adversarial}. Some of these works account for interference between units, but they typically assume that the interference is either explicitly known to the learning algorithm, or the interference follows a specific pattern. In contrast, our setting makes no such assumptions about the nature or structure of interference between units.

Our paper is closely related to the field of multitasking bandits, where the learning algorithm is designed to achieve multiple objectives simultaneously during the learning process. \citet{yang2017framework} explore the regulation of the false discovery rate while identifying the best arm. \citet{yao2021power} focus on ensuring the ability to detect whether an intervention has an effect, while also leveraging contextual bandits to tailor consumer actions. \citet{Jamieson2013lilU,cho2024reward} aim to minimize cumulative regret while identifying the best arm with minimal sample complexity. 
\citet{erraqabi2017trading} aims to balance the trade-off between regret minimization and estimation error; however, their design can not guarantee optimality.

% \paragraph{Multitasking bandits}
% \paragraph{Bandit with inference}

\begin{table}[t]
\scalebox{0.8}{
\begin{tabular}{cccc}
 \toprule[1.5pt]
    & Estimation (offline)  & Regret (online) & Trade-off between Estimation\&Regret \\
    \hline
    Without interference & SUTVA causality &  {\makecell[c]{ \citet{Auer2002FinitetimeAO}\\ \citet{burtini2015survey}}} & {\makecell[c]{\citet{simchi2023multi} \\ \citet{duan2024regret} }}\\
    \hline
     With interference & {\makecell[c]{{\citet{leung2022causal, leung2022rate, leung2023network}} \\  \citet{hudgens2008toward}\\ \citet{savje2024causal}}}& \makecell[c]{\citet{agarwal2024multi} \\ \citet{jia2024multi} \\ \citet{xu2024linear}} & \textbf{Our paper}\\
      \toprule[1.5pt]
\end{tabular} }
\caption{Most related and representative works in causality estimation and regret analysis with (without) network interference. }\label{tab_compare}
\end{table}

\paragraph{Trade-off between inference (estimation) and regret.}
A significant body of research has been dedicated to developing statistical methods for inference in MABs. Numerous studies focus on deriving statistical tests or central limit theorems for MABs while ensuring that the bandit algorithm remains largely unaltered \citep{hadad2021confidence,luedtke2016statistical,deshpande2023online,zhang2020inference,zhang2021statistical,han2022online,dimakopoulou2017estimation,dimakopoulou2019balanced,dimakopoulou2021online}, thereby facilitating aggressive regret minimization. However, these works all rely on the SUTVA assumption and fail to account for potential interference between units.

Previous literature upon adaptive inference in multi-armed bandits include~\citet {dimakopoulou2021online, liang2023experimental} whereas without strict trade-off analysis. To our best knowledge, the only state-of-the-art trade-off result is primarily constructed by~\citet {simchi2023multi} whereas also be cursed by the SUTVA assumption without a network connection. Moreover, \citet{duan2024regret} argue that such Pareto-optimality could be further improved, i.e., the regret and estimation error could simultaneously achieve their optimality, if additionally assuming the ``covariate diversity'' of each node without network interference. Stepping forward, when we shift our attention to the network setting, \citet{jia2024multi} is also intuitively aware of the potential “incompatibility” of decision-making and statistical inference: specifically, \citet{jia2024multi} emphasizes that the truncated HT estimator directly into the policy learning system is no longer robust because policy learning gives different propensity probabilities to different arms, making the propensity score more extreme. 

\section{Justification, discussion and future work}\label{app_discussion}
\paragraph{Justification on exposure mapping.} It is a well-known concept in causality. From a statistical perspective, it serves as a functional tool for mapping a high-dimensional action space to a low-dimensional manifold; from a machine learning standpoint, it can be interpreted as a specialized input representation layer.
However, its utility has not been fully explored in interference-based online learning settings like Bandits. Interference-based bandit referred to as exposure mapping has recently been explored in~\citet{jia2024multi} to our knowledge. This additionally assumes the intensity of interference decays with distance. Still, the low-dimensional vectors from their exposure mapping are not involved in the computation of the target regret. In contrast, their regret, directly uses the adversarial setting that “the original super arm must be a vector of the form $a * \bold{1}^N, a \in \mathcal{K}$”, which is limited in realistic compared to our settings, e.g. when the optimal arm takes place when the individuals in the network are assigned to different treatments; to tackle this problem, although \citet{agarwal2024multi} can identify the best arm beyond $a*\mathbf{1}^N, a \in \mathcal{K}$, their approach relies on a stronger assumption: the rewards of each node are influenced solely by its limited first-order neighbors, and the number of these neighbors is significantly smaller than $N$.
In sum, our paper first presents an integration of exposure mapping with bandit regret frameworks and demonstrates its generality and applicability.

% In comparison, exposure mapping can incorporate more generalized information by definition; in addition, another series of powerful tools for reducing action space is clustering, which we refer the reader to~\citet{zhang2023individualized, viviano2023causal} and will continue to discuss later.

% \paragraph{Justification on noise assumption} In our work, we assume that the noise follows an arbitrary zero-mean i.i.d 1-sub gaussian distribution which is bounded by. It is mild, following~\citet{agarwal2024multi}, and is more general than the boundedness assumption~\citet{simchi2023multi}, whereas our trade-off and optimality result holds the same and also with detailed algorithms. Moreover, we further discuss the design-based setting in the discussion part and make a comparison with~\citet{jia2024multi}'s setting,  where all randomness is generated from the treatment assignment, and all other values are fixed.

% \subsection{Extension on arm size: when $K > 2$}

\paragraph{Justification on Condition~\ref{condition_in}.} Condition~\ref{condition_in} states that $\mathcal{U}_{\mathcal{E}} \geq 2$ is not empty. It is already weaker than the previous interference-based bandit setting \citep{jia2024multi, agarwal2024multi} whereas it could be further relaxed. We consider the generalized metric to describe the distance between $\mathcal{U}_{\mathcal{O}}$ and $S_t \in \mathcal{U}_{\mathcal{C}}$:  $\mathcal{D}(\mathcal{U}_{\mathcal{O}}, S_t):= \min_{ S'\in \mathcal{U}_{\mathcal{O}} }||S'-S_t||_1$ via Manhattan distance. When the number of clusters grows, the action space $|\U_{\mathcal{C}}|$ exponentially expands and their compatibility $\mathcal{D}(\mathcal{U}_{\mathcal{C}},  \mathcal{U}_{\mathcal{O}} ) $ also decreases. These previous literature and Condition~\ref{condition_in} all satisfy $\mathcal{D}(\mathcal{U}_{\mathcal{C}}, \mathcal{U}_{\mathcal{O}}) = 0$, and the former literature  together with additional network structure~\citep{Agrawal2012ThompsonSF} or interference intensity~\citep{jia2024multi} assumption as above. In Appendix~\ref{optimize_perspective} we claim that under the weakened assumption $\mathcal{D}(\U_{\cC}, \U_{\cO}) \leq \epsilon$, where $\epsilon>0$ is a prior constant, our model remains capable of reasonable modeling by appropriately adjusting the definition of exposure-based rewards accordingly. The interplay between this assumption and other well-known assumptions, such as the neighbor sparsity assumption~\citet{agarwal2024multi}, the decaying interference assumption~\citet{jia2024multi}, and the approximate interference assumption~\citet{leung2022causal}, is left as an avenue for future work.

\section{Experiments}\label{ex_app}

\begin{figure*}[ht]
\centering
\includegraphics[width=0.45\linewidth]{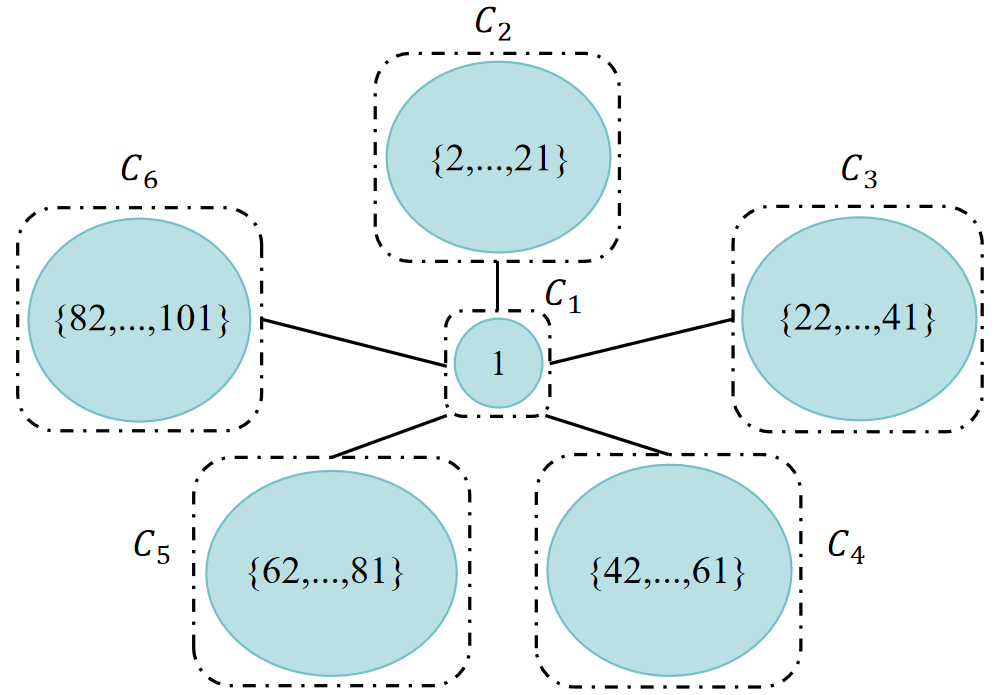}
\caption{Network structure.}\label{fig.topo}
\end{figure*}

\begin{figure*}[ht]
\centering
	\subfigure[Cumulative regret]{
\quad\includegraphics[width=0.36\linewidth]{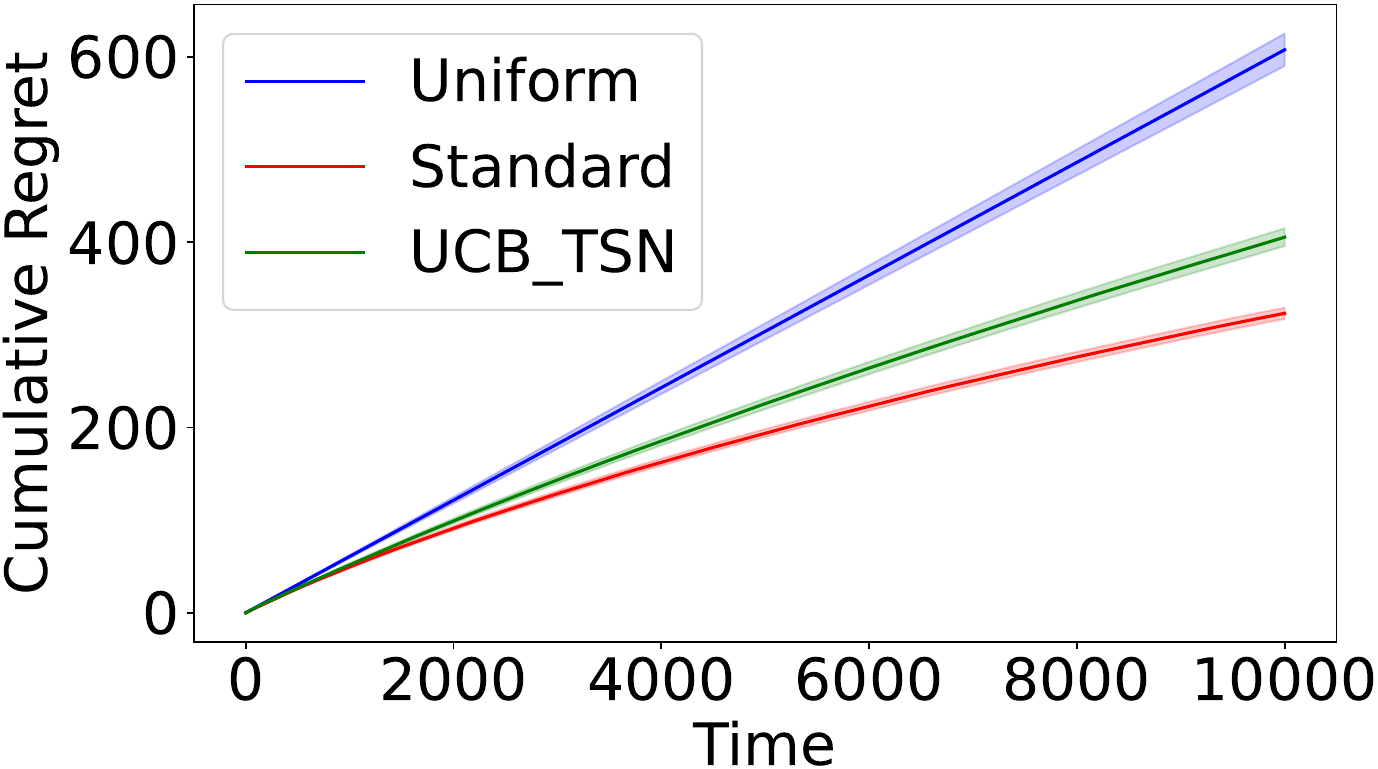}\label{fig2b}}
\subfigure[ATE estimation error]{
\includegraphics[width=0.56\linewidth]{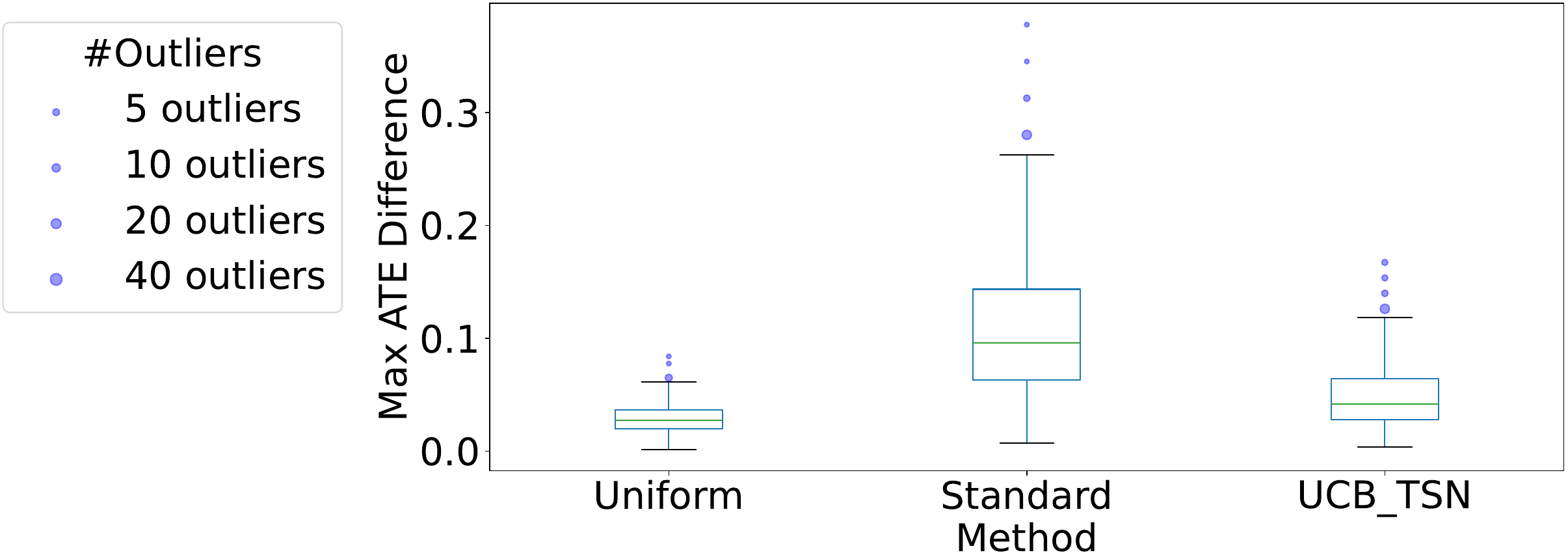}\label{fig2c}}
\caption{Experimental results.}
\end{figure*}
% \begin{figure}
%     \centering
%     \includegraphics[width=0.5\linewidth]{}
%     \caption{Caption}
%     \label{fig:enter-label}
% \end{figure}
\paragraph{Setup.} We consider a network consisting of 101 units. Specifically, there is a central cluster \(C_1 = \{1\}\) that contains a single unit, which is connected to every unit in the five peripheral clusters \(C_2, \dots, C_6\) (namely, \(C_2 = \{2, \dots, 21\}\), \(C_3 = \{22, \dots, 41\}\), \(C_4 = \{42, \dots, 61\}\), \(C_5 = \{62, \dots, 81\}\), and \(C_6 = \{82, \dots, 101\}\), with each outer cluster containing 20 units, as shown in Fig. \ref{fig.topo}). We set the action set as \(\mathcal{K} = \{0,1\}\). Inspired by \citep{leung2022causal,gao2023causal}, we define the exposure mapping as \(\bS(i, A, \bH) = \mathbf{1}\!\left\{\frac{\sum_j h_{ij} a_j}{\sum_j h_{ij}} \in \left[0, \frac{1}{2}\right)\right\}\), which explores the influence of the proportion of neighbors taking action \(1\) on each unit; this exposure mapping implies that \(d_s = 2\). For every \(S \in \mathcal{U}_{\bbE}\), we define \(\bP(A_t = A \mid S)\) as uniform sampling. Moreover, for each selected super arm corresponding to an exposure $S$, the reward is sampled from a Bernoulli distribution.

We evaluate the performance of \texttt{UCB-TSN} ($T_1 = \sqrt{|\mathcal{U}_{\bbE}|\, T}$) against two baseline methods: Standard (i.e., \texttt{UCB-TSN} with \(T_1 = 0\)) and Uniform (i.e., \texttt{UCB-TSN} with \(T_1 = T\)). Each algorithm is executed 1000 times, and we report the averaged results. 

\paragraph{Results.} The simulation results are shown in Fig.~\ref{fig2b} and Fig.~\ref{fig2c}. As seen in Fig.~\ref{fig2b}, both the Standard method and \texttt{UCB-TSN} achieve the lowest cumulative regret, while Uniform exhibits the highest cumulative regret. Fig.~\ref{fig2c} presents a box plot of the maximum ATE estimation error, \(e_\nu(T,\hat{\Delta})\), where the green line represents the median. The results indicate that \texttt{UCB-TSN} and Uniform yield lower ATE estimation errors with compact interquartile ranges and few outliers, whereas the Standard method shows a wider spread of errors and multiple outliers. This relatively poorer performance of the Standard method in statistical estimation is due to its lower frequency of exploring sub-optimal arms compared to Uniform and \texttt{UCB-TSN}. 

\section{The Discussion of Exposure Mapping and Noise Rescaling}\label{app_exposure_sec}

We denote the policy and exposure reward inheriting from~\citet{leung2022causal} as $\mathbb{P}_{\text{Leung}}$ and $\tilde{Y}_{i,\text{Leung}}(\cdot)$, respectively. Considering Eq~\eqref{exposure}, we take the exposure mapping function's output as $d_s$ cardinality without loss of generality. We choose $\mathbb{P}(A_t = A\mid S_t) := \mathbb{P}_{\text{Leung}}(A_t = A \mid S_t \bm{e}_i)$ then $\forall  S_t \bm{e}_i= s, \tilde{Y}_{i,\text{Leung}}(s) = \sum_{A\in\mathcal{K}^\U} \mathbb{P}_{\text{Leung}}(A_t = A \mid s)Y_i(A) = \sum_{A\in\mathcal{K}^\U} \mathbb{P}_{\text{}}(A_t = A \mid S_t)Y_i(A) = \tilde{Y}_i(S_t)$. Hence our exposure-based reward notation is generalized from~\citet{leung2022causal}.

Moreover, we discuss the re-scaling of noise. When $\forall S \in \mathcal{U}_{\mathcal{E}}, |\{A: \{\bS(i, A, \mathbb{H})\}_{i \in \U} = S\}| = 1$, it naturally leads to the variance proxy $\sigma^2 = \frac{1}{N}$ of the Sub-Gaussian variables $\sum_{i \in \mathcal{U}} \tilde{r}_{i,t}(S)/N$. Hence, we mainly consider other cases. Notice that Eq~\eqref{exposure} defines 
\begin{equation*}
\begin{aligned}
[\Tilde{Y}_i({S}_{t}), \Tilde{r}_{i,t}({S}_{t})]^{\top} := \sum_{A \in \mathcal{K}^{\U}} [Y_i(A), {r}_{i,t}(A)]^{\top} \mathbb{P}_{}(A_t =A \mid {S}_t), 
\end{aligned}
\end{equation*}
namely, for each $S_t$, practitioners select random legitimate $r_{i,t}(A_t)$ to approximate $\tilde{r}_{i,t}(S_t)$, each with probability $\mathbb{P}(A_t =A \mid S_t )$. The randomness includes the sub-Gaussian noise and sampling noise. It follows that for all $m\in\mathbb{R}$,
\begin{equation}
\begin{aligned}
&\mathbb{E}\Bigg[\text{exp}\bigg({\frac{m}{N}\sum_{i\in \mathcal{U}}\Big(\Tilde{r}_{i,t}({S}_{t}) - \Tilde{Y}_{i}({S}_{t})\Big)}\bigg) \mid A_t = A\Bigg]\\ =
&
\mathbb{E}\Bigg[\text{exp}\bigg({\frac{m}{N}\sum_{i\in \mathcal{U}}\Big({r}_{i,t}({A}_{}) - {Y}_{i}({A}_{}) + {Y}_{i}({A}_{}) - \Tilde{Y}_{i}({S}_{t})\Big)}\bigg) \Bigg] \\
=& \text{exp}\bigg({\frac{m}{N}\sum_{i\in \mathcal{U}}\Big({Y}_{i}({A}_{}) - \Tilde{Y}_{i}({S}_{t})\Big)}\bigg)\mathbb{E}\Bigg[\text{exp}\bigg({\frac{m}{N}\sum_{i\in \mathcal{U}}\Big({r}_{i,t}({A}_{}) - {Y}_{i}({A}_{})\Big)}\bigg) \Bigg] \\
\leq & \text{exp}\bigg({\frac{m}{N}\sum_{i\in \mathcal{U}}\Big({Y}_{i}({A}_{}) - \Tilde{Y}_{i}({S}_{t})\Big)}\bigg) \text{exp}\Big({\frac{m^2}{2N}}\Big).
\end{aligned}\label{app_exposure}
\end{equation}

Taking expectation upon both sides of Eq~\eqref{app_exposure}, it leads to
\begin{equation}
\begin{aligned}
\mathbb{E}\Bigg[\mathbb{E}\Bigg[\text{exp}\bigg({\frac{m}{N}\sum_{i\in \mathcal{U}}\Big(\Tilde{r}_{i,t}({S}_{t}) - \Tilde{Y}_{i}({S}_{t})\Big)}\bigg) \mid A_t = A\Bigg]\Bigg] \leq \text{exp}\Big({\frac{m^2}{2N}}\Big)\mathbb{E}\Bigg[ \text{exp}\bigg({\frac{m}{N}\sum_{i\in \mathcal{U}}\Big({Y}_{i}({A}_{}) - \Tilde{Y}_{i}({S}_{t})\Big)}\bigg)\Bigg]. 
\end{aligned}\label{app_exposure_1}
\end{equation}
According to the boundary $\frac{1}{N}\sum_{i\in \mathcal{U}}({Y}_{i}({A}_{}) - \Tilde{Y}_{i}({S}_{t})) \in [-1,1]$, it is natural to derive 
\begin{align}
\nonumber
\mathbb{E}\Bigg[ \text{exp}\bigg({\frac{m}{N}\sum_{i\in \mathcal{U}}\Big({Y}_{i}({A}_{}) - \Tilde{Y}_{i}({S}_{t})\Big)}\bigg) \Bigg] \leq \text{cosh}(m/2)\leq \text{exp}(m^2/8).
\end{align}
% According to the Hoeffding's inequality, it achieves that $\mathbb{E} \text{exp}({m\frac{1}{N}\sum_{i\in \mathcal{U}}({Y}_{i}({A}_{}) - \Tilde{Y}_{i}({S}_{t}))}) \leq \text{exp}(m^2/2)$. 
Then Eq~\eqref{app_exposure_1} achieves that
\begin{equation}
\begin{aligned}
\eqref{app_exposure_1} \leq \text{exp}\Big({\frac{m^2}{2N}}\Big) \text{exp}(m^2/8) = \text{exp}\bigg(\frac{m^2}{2}\Big(\frac{1}{N}+\frac{1}{4}\Big)\bigg).
\end{aligned}
\end{equation}
Therefore the Sub-Gaussian variables $\sum_{i \in \mathcal{U}} \tilde{r}_{i,t}(S)/N$ could achieve the variance proxy at most ${1}/{N}+{1}/{4}$. In the following part, we set the variance proxy as $\sigma^2 = 2$ without loss of generality.

\paragraph{Comment on the order of node number $N$.} For a supplement, in Theorem/Corollary~\ref{stochasticATE}-\ref{trade-off1},  we additionally consider the order of node number $N$. (i) In Theorem~\ref{stochasticATE}, we emphasize that if $\forall S' \in \mathcal{U}_{\mathcal{E}}, |\{A: \{S(i, A, \mathbb{H})\}_{i \in \U} = S'\}| = 1$, namely, there is only one legitimate $A$ which is compatible with each exposure arm $S'$, then Theorem~\ref{stochasticATE} could be strengthened as $\bE\big[\vert \hat{\Delta}_T^{(i,j)} - \Delta^{(i,j)} \vert\big] = \tilde{O}\big(\sqrt{{\vert \U_\bbE \vert}/{T_1}N}\big)$. Take the cluster-wise switchback experiment ($S(i, A, \mathbb{H}) =a_{i,t}$) for instance, which is the generalized case of~\citet{jia2024multi}. In this case, since $|\mathcal{U}_{\mathcal{E}}| = K^C \ll N$ via manually selecting $d_s, C$, then we can claim the estimation is consistent when $N \rightarrow +\infty$\footnote{Essentially, it is due to the re-scaling of noise. Under the one-to-one mapping in this paragraph, the result is intuitive since $\sum_{i \in \mathcal{U}} \tilde{r}_{i,t}(S)/N$ exhibits a re-scaled Sub-Gaussian noise with variance proxy $1/N$. It degenerates to the offline setting when $N \rightarrow +\infty$. Otherwise, we could only ensure $\sum_{i \in \mathcal{U}} \tilde{r}_{i,t}(S)/N$ is a Sub-Gaussian noise with variance proxy $({1}/{N}+{1}/{4})$. We defer the details to Appendix~\ref{app_exposure_sec}.}. Moreover, in the setting of~\citet{agarwal2024multi}, it is equivalent to the case $C= N$ and thus the result in Theorem~\ref{stochasticATE} is transformed as $\tilde{O}\big(\sqrt{{K^N}/{T_1 N}}\big)$. It serves as a supplement of Proposition~\ref{negative}, claiming that not only the regret but also the estimation error is hard to control without exposure mapping. (ii) Analogously, in Theorem~\ref{stochasticregret}, the result is transformed to $\bR_{\nu}(T,\pi) = \tilde{O}\big(\sqrt{\vert \U_\bbE \vert T/N} + T_1\big)$ under the above one-to-one mapping. (iii) Finally, in Corollary~\ref{trade-off1}, the trade-off is transferred to be $\tilde{O}(\sqrt{\vert \U_\bbE\vert/N })$ when we slightly modify the condition of $T_1$ as $T_1 \geq \sqrt{\vert \U_\bbE \vert T/N} \vee |\mathcal{U}_{\mathcal{E}}|$. This result is also aligned with the proof of Theorem~\ref{trade-off}.

% \zhiheng{better THAN chonghuan choose N=1N=1 and |UE|=K|\mathcal{U}_{\mathcal{E}}| = K. }

\section{Proof of Proposition~\ref{negative}}\label{app_propostion1}

\begin{proof}[Proof of Proposition~\ref{negative}] We here define $\bK^\U := \{A_k\}_{k=1}^{K^N}$ as the set of the super arm. Define a MAB instance $\nu_1 \in \bbE_0$ that $Y_i(A) = \Delta \bone\{ A = A_1 \}$ for all $i \in \U$ and $A\in\bK^\U$, where $\Delta \in [0,1/2]$ will be defined later. We suppose that the noise of all unit $\eta_{i,t}$ follows a $\mathcal{N}(0,1)$ Gaussian distribution, and therefore the normalized noise of the super arm $(1/N)\sum_{i\in\U} \eta_{i,t}$ follows a $\mathcal{N}(0,1/N)$ Gaussian distribution.  Hence, we have $1/N \sum_{i\in\U} Y_i(A_1) = \Delta$ and $1/N \sum_{i\in\U} Y_i(A_k) = 0$ for all $k\in [K^N]/\{1\}$. This implies in $\nu_1$, $A_1 = A^*$ is the best arm with potential outcome $\Delta$ and $A \not = A_1$ is the sub-optimal arm with potential outcome $0$. Due to 
\begin{align}\label{decomregret}
        \bR_{\nu_1}(T,\pi) = \sum_{k=2}^{K^N} \Delta_k \bE_{\nu_1,\pi}[\N_{A_k}^T],
    \end{align}
    where $\N_{A_k}^T := \sum_{t \in [T]} \bone\{A_t = A_k\}$ denotes the number that supper arm $A_k$ is trigger till $T$ and $\Delta_k$ denotes the reward gap between super arm $A_1$ and $A_k$ (i.e., $\Delta_k = (1/N) (\sum_{i\in\U} Y_i(A_1) - Y_i(A_k) )$). Suppose the super arm $A_j$, $j = \arg\min_{j \in [K^N]/\{1\}} \bE_{\nu_1,\pi}[\N_{A_j}^T] $, then
    \begin{align}
        \bE_{\nu_1,\pi}[\N_{A_j}^T] \le \frac{T}{K^N - 1}.
    \end{align}
    Besides, we define another $\mathcal{N}(0,1)$ Gaussian MAB instance $\nu_2 \in \bbE_0$, where $Y^\prime_i(A) = Y_i(A) + 2\Delta \bone\{A = A_j\}$ for all $i\in\U$ and $A \in \bK^\U$. In $\nu_2$, $A_j$ is the best arm with potential outcome $2\Delta$. Based on the decomposition of the regret Eq~\eqref{decomregret}, we have
    \begin{align}
        \bR_{\nu_1}(T,\pi) \ge \bP_{\nu_1,\pi}\big(\N_{A_1}^T\le T/2\big)\frac{\Delta T}{2},\quad\text{and}\quad\bR_{\nu_2}(T,\pi) \ge \bP_{\nu_2,\pi}\big(\N_{A_1}^T \ge T/2\big)\frac{\Delta T}{2}.
    \end{align}
 
Let $\mathbb{P}_{\nu_1,\pi}$ and $\mathbb{P}_{\nu_2,\pi}$ denote the probability measures on the canonical bandit model induced by the $T$-round interaction between $\pi$ and $\nu_1$, and $\pi$ and $\nu_2$, respectively. Finally, we have
    \begin{align}
    \begin{split}
&\bR_{\nu_1}(T,\pi) + \bR_{\nu_2}(T,\pi) \\ \ge &  \Big(\bP_{\nu_1,\pi}\big(\mathcal{N}_{A_1}^T \ge T/2\big) + \bP_{\nu_2,\pi}\big(\mathcal{N}_{A_1}^T < T/2\big)\Big)\frac{\Delta T}{2}\\
\ge & \text{exp}\Big(-\text{KL}(\mathbb{P}_{\nu_1,\pi},\mathbb{P}_{\nu_2,\pi})\Big)\frac{\Delta T}{4}\\
\ge & \text{exp}\bigg( -  \bE_{\nu_1,\pi}[\N_{A_j}^T] \text{KL}\Big(\mathcal{N}(0,1/N),\mathcal{N}(2\Delta,1/N)\Big)\bigg) \frac{\Delta T}{4} \\
\ge & \text{exp}\bigg( - \bE_{\nu_1,\pi}[\N_{A_j}^T]  2N\Delta^2\bigg) \frac{\Delta T}{4}\\
\ge & \text{exp}\bigg( -\frac{2TN\Delta^2}{K^N-1} \bigg) \frac{\Delta T}{4},
    \end{split}
    \end{align}
    where $\text{KL}$ denotes the KL divergence, the second inequality is owing to the Bretagnolle–Huber inequality, the third inequality is due to the Lemma 15.1 in \citet{lattimore2020bandit}, the fourth inequality is due to the definition of the noise distribution (i.e., $\mathcal{N}(0,1/N)$) of the super arm. Finally, select $\Delta =\sqrt{\frac{K^N-1}{4TN}}  \wedge \frac{1}{2}$, based on the above result, we have $(i = 1\text{~or~}2)$
    \begin{align}
\bR_{\nu_i}(T,\pi) \ge \begin{cases}e^{-1/2}\frac{T}{8\sqrt{N}},\text{~when~} T \leq K^N\\
    \frac{e^{-1/2}}{4} \sqrt{\frac{(K^N-1)T}{N}},\text{~when~} T \geq K^N.
\end{cases} 
    \end{align}
\end{proof}

\section{Proof of Theorem~\ref{trade-off}}\label{app_tradeoff}

% \begin{lemma}\label{prepare}
% Given $\mathbf{S}$ and $\mathcal{C}$ that satisfies Condition~\eqref{condition_in}. Given any online decision-making policy $\pi$, any ATE estimator has a lower-bounded probability error as follows:

% \begin{equation}
% \begin{aligned}
% \inf _{\hat{\Delta}_T} \max _{\nu \in \mathcal{E}_0} \mathbb{P}_\nu\left(\left|\hat{\Delta}_T-\Delta_\nu\right| \geq \phi(T)\right) \geq \frac{1}{2}\left[1-\sqrt{g (\phi(T)) \frac{\mathcal{R}_\nu(T, \pi)}{\left|\Delta_\nu\right|}}\right] .
% \end{aligned}
% \end{equation}
% \end{lemma}

    % The sketch of proof: 
    % \begin{itemize}
\begin{proof}[Proof of Theorem~\ref{trade-off}]
        In this section, to simplify the notations in Section \ref{app_propostion1}, we abbreviate  $\mathbb{P}_{\nu,\pi}$ as $\mathbb{P}_\nu$ and $\mathbb{E}_{\nu,\pi}$ as $\mathbb{E}_{\nu}$.
        We consider two kinds of instances for a fixed policy $\pi$ and a fixed strategy of constructing an ATE estimator $\hat{\Delta}_T$. For the first one (i.e., $\nu_1$), we denote it as $r_{i,t}(A) = f_i(A) + \eta_{i,t} $. Here we let $Y_i(A) := f_i(A) \in (\varepsilon_0,1-\varepsilon_0), \varepsilon_0 \in (0,1), r_{i,t}(A) \in \{-1,1\}$. It means
         $r_{i,t}(A) = \text{Rad}(\frac{1-f_i(A)}{2}, \frac{1+f_i(A)}{2})$. For each feasible cluster-wise super exposure arm $S \in \U_\bbE$, recall that
        \begin{equation}
        \begin{aligned}
        \Tilde{Y}_i(S_{}) &= \sum_{A \in \mathcal{K}^{\U}} f_i(A) \mathbb{P}_{}(A_t = A \mid S).
        \end{aligned}
        \end{equation}
The difference of expected reward of $S,S'$ could be represented by $\Delta_1(S,S') := \frac{1}{N}\sum_{i \in \mathcal{U}} (\Tilde{Y}_{i}(S ) - \Tilde{Y}_{i}(S'  ))$, which is 
\begin{equation}
\begin{aligned}
  &\Delta_1(S, S') = \frac{1}{N}\sum_{i \in \mathcal{U}} \sum_{A \in \mathcal{K}^{\U}} f_i(A) \big(\mathbb{P}_{}(A_t = A\mid S )  
-  \mathbb{P}_{}(A_t = A \mid S' ) \big).
\end{aligned}\label{delta_1}
\end{equation}

% For brevity, we denote pπ(i,At,S):=Pπ(At|S(i,At,H)=S)p_{\pi}(i,A_t, {S}) := \mathbb{P}_{\pi}(A_t|\textbf{S}(i,A_t,\mathbb{H})=S ), which is analogous to the case of S′S'. It implies that ATE between arm S,S′S, S' in such instance is
% \begin{equation}
%     \begin{aligned}
%     \Delta_1 := \frac{1}{N}\sum_{i \in \mathcal{U}} \sum_{A_t \in \mathcal{K}^{\U}} f_i(A_t) (p_{\pi}(i, A_t,S) - p_{\pi}(i, A_t,S')).
%     \end{aligned}\label{ate_1}
% \end{equation}

Without loss of generality, we select the feasible super arm to set $\Delta_1(S, S')< 0$. For brevity, we omit the expression of the parentheses in the following text. Namely, we choose $S'$ as the best arm, and $S$ as a sub-optimal arm in $\U_\bbE$. We choose $S = \arg\min_{S_i\in\U_\bbE} \Delta_1(S_i,S') \mathbb{E}_{\nu_1}[\mathcal{N}_{S_i}^T]$. In this process, we use $\hat{\Delta}^{(i,j)} := \{\hat{\Delta}^{(i,j)}_t\}_{t \geq 1}, \hat{\Delta} := \{\hat{\Delta}^{(i,j)}\}_{S_i,S_j \in \mathcal{U}_{\mathcal{E}}}$. We then construct a new MAB instance $\nu_2$ and hope to get a different ATE value. We define it as $r'_{i,t}(A)$. We establish :
        \begin{equation}
        \begin{aligned}
          r'_{i,t}(A) := \begin{cases}
                r_{i,t}(A) &\forall A\text{~satisfying~} \mathbb{P}_{}(A_t = A \mid S ) = 0.   \\
               \text{Rad}(\frac{1-f_i(A)+\alpha}{2}, \frac{1+f_i(A)-\alpha}{2}) & \forall A\text{~satisfying~} \mathbb{P}_{}(A_t = A \mid S )  > 0.
          \end{cases}
        \end{aligned}\label{construct}
        \end{equation}
Here $\alpha>0$ could be chosen sufficiently small. Remind that following Eq~\eqref{delta_1}, the ATE between super arm $S,S'$ is
\begin{equation}
\begin{aligned}
\nonumber
 \Delta_2 :&= \Delta_{2,1}+ \Delta_{2,2},~\text{where~}\\
 \Delta_{2,1} &:= \frac{1}{N}\sum_{i \in \mathcal{U}} \sum_{A \in \mathcal{K}^{\U}}   (f_i(A)-\alpha)  \big(\mathbb{P}_{}(A_t = A \mid S )  - \mathbb{P}_{}(A_t = A\mid S' ) \big)\bone\{\mathbb{P}_{}(A_t = A \mid S ) >0 \},\\
  \Delta_{2,2} &:= \frac{1}{N}\sum_{i \in \mathcal{U}} \sum_{A \in \mathcal{K}^{\U}} {f_i(A)} \big(\mathbb{P}_{}(A_t = A \mid S )  - \mathbb{P}_{}(A_t = A \mid S' ) \big)\bone\{\mathbb{P}_{}(A_t = A \mid S ) =0\}.
\end{aligned}
\end{equation}
        Hence, it implies that the ATEs in these two MAB instances, respectively, contain a difference
        \begin{equation}
            \begin{aligned}
                &\Delta_2 - \Delta_1 \\=& \frac{1}{N}\sum_{i \in \mathcal{U}} \sum_{A \in \mathcal{K}^{\U}} -\alpha \big(\mathbb{P}_{}(A_t = A \mid S )  -\mathbb{P}_{}(A_t = A \mid S' ) \big)\bone\{\mathbb{P}_{}(A_t = A \mid S ) >0\}\\
                =& \frac{1}{N}\sum_{i \in \mathcal{U}} \sum_{A \in \mathcal{K}^{\U}} -\alpha \mathbb{P}_{}(A_t = A\mid S )  \bone\{\mathbb{P}_{}(A_t = A \mid S ) >0\}\\
                =& \frac{1}{N}\sum_{i \in \mathcal{U}} \sum_{A \in \mathcal{K}^{\U}} -\alpha \mathbb{P}_{}(A_t = A \mid S )  = -\alpha < 0.
            \end{aligned}
        \end{equation}
        Naturally, our setting leads to $0 > \Delta_1 > \Delta_2$. The second equality is because $\mathbb{P}_{}(A_t = A \mid S ) \mathbb{P}_{}(A_t = A\mid S' )  = 0$ when $S \neq S'$. In this sense, we consider a given estimate strategy, which is summarized by $\{\hat{\Delta}_{t'}\}_{ t' \in [t]}$. We define a minimum test $\psi(\hat{\Delta}_t) = \arg\min_{i=\{1,2\}}|\hat{\Delta}_t - \Delta_i|$. Naturally, it implies that $\psi(\hat{\Delta}_t) \neq i, i\in \{1,2\}$ is a sufficient condition of $|\hat{\Delta}_t - \Delta_i| \geq \frac{\alpha}{2} $. As a consequence,
    \begin{equation}
\begin{aligned}
\inf _{\hat{\Delta}_t} \max _{\nu \in \mathcal{E}_0} \mathbb{P}_\nu\left(|\hat{\Delta}_t-\Delta_\nu| \geq \frac{\alpha}{2} \right) & \geq \inf _{\hat{\Delta}_t} \max _{i \in\{1,2\}} \mathbb{P}_{\nu_i}\left(|\hat{\Delta}_t-\Delta_i| \geq \frac{\alpha}{2}\right) \\
& \geq \inf _{\hat{\Delta}_t} \max _{i \in\{1,2\}} \mathbb{P}_{\nu_i}\left(\psi(\hat{\Delta}_t ) \neq i\right) \\
& \geq \inf _\psi \max _{i \in\{1,2\}} \mathbb{P}_{\nu_i}(\psi \neq i).
\end{aligned}\label{min_max_1}
\end{equation}

Here, the probability space is constructed on the exposure arm $\{\textbf{S}(i, A,\mathbb{H})\}_{i \in \mathcal{U}}$ in each time period $t$, and the observed exposure reward. 
% is constructed upon the observations that have been collected, namely, the set of {ai,t,ri,t}\{a_{i,t}, r_{i,t}\}, where i∈[N],t=1,2,...T.i \in [N], t = 1,2,...T. Moreover, ψ\psi represents all kinds of hypothetical tests constructed by {ai,t,ri,t}i∈[N],t∈[T]\{a_{i,t}, r_{i,t}\}_{i \in [N], t \in [T]}. Moreover, ˆΔt\hat{\Delta}_t denotes any pre-specified estimator constructed by {ai,t,ri,t}i∈[N],t∈[T]\{a_{i,t}, r_{i,t}\}_{i \in [N], t \in [T]}.
We use the technique in min-max bound. Notice that the original feasible region of MAB instances as $\mathcal{E}_{0}$; we get  
\begin{equation}
\begin{aligned}
\text{RHS~of~}~\eqref{min_max_1} \geq& \inf_{\psi}\max_{i \in \{1,2\}} \mathbb{P}_{\nu_i}(\psi \neq i) \\
\geq & \frac{1}{2} \inf_{\psi}(\mathbb{P}_{\nu_1}(\psi = 2) + \mathbb{P}_{\nu_2}(\psi = 1))\\
= & \frac{1}{2}(1-\text{TV}(\mathbb{P}_{\nu_1}, \mathbb{P}_{\nu_2}))\\
\geq & \frac{1}{2}\bigg[1-\sqrt{\frac{1}{2}\text{KL}(\mathbb{P}_{\nu_1}, \mathbb{P}_{\nu_2})}\bigg].
\end{aligned}\label{min_max_2}
\end{equation}
We aim to provide an upper bound of KL divergence $\text{KL}(\mathbb{P}_{\nu_1}, \mathbb{P}_{\nu_2})$, inspired by the divergence decomposition:

\begin{equation}
\begin{aligned}
\text{KL}(\mathbb{P}_{\nu_1}, \mathbb{P}_{\nu_2}) = \mathbb{E}_{\nu_1} \left[\log \left(\frac{\mathrm{d} \mathbb{P}_{\nu_1}}{\mathrm{d} \mathbb{P}_{\nu_2^{}}}\right)\right].
\end{aligned} \label{kl}
\end{equation}

For any instance $\nu \in \{\nu_1,\nu_2\}$, the density function of the series is denoted as (we denote $X_t$ as the observed exposure reward $\{ \tilde{r}_{i,t}(S) \}_{i\in\U}$)
\begin{equation}
\begin{aligned}
\mathbb{P}_{\nu}\left(S_1, X_1, \ldots, S_t, X_t \right) = \prod_{t'=1}^t \pi_t\left(S_t \mid S_1, X_1, \ldots, S_{t'-1}, X_{t'-1} \right) \mathbb{P}_{\nu, S_t}\left(X_t\right).
\end{aligned}
\end{equation}

Here $\mathbb{P}_{\nu, S}(\cdot)$ denotes the reward density distribution conditioning on arm $S$ in $\nu$. Hence Eq~\eqref{kl} can be transformed as
\begin{equation}
    \begin{aligned}
    \text{KL}(\mathbb{P}_{\nu_1}, \mathbb{P}_{\nu_2}) =& \sum_{t' \in [t]} \mathbb{E}_{\nu_1} log\Big(\frac{\mathbb{P}_{\nu_1, S_{t'}}(X_{t'})}{\mathbb{P}_{\nu_2, S_{t'}}(X_{t'})}\Big)\\
    =& \sum_{t' \in [t]} \mathbb{E}_{\nu_1} \Big[\mathbb{E}_{\nu_1} log\Big(\frac{\mathbb{P}_{\nu_1, S_{t'}}(X_{t'})}{\mathbb{P}_{\nu_2, S_{t'}}(X_{t'})}\Big) \mid S_{t'} \Big] \\
    =& \sum_{t' \in [t]} \mathbb{E}_{\nu_1} \big[\text{KL}( \mathbb{P}_{\nu_1, S_{t'}}(\cdot), \mathbb{P}_{\nu_2, S_{t'}}(\cdot))\big] \\
    =& \mathbb{E}_{\nu_1}[\N_{S}^t] \text{KL}( \mathbb{P}_{\nu_1, S}(\cdot), \mathbb{P}_{\nu_2,S}(\cdot)).
    \end{aligned}
\end{equation}
The last equation is derived from the construction in Eq~\eqref{construct}. We aim to compute $\text{KL}(\mathbb{P}_{\nu_1,S}(\cdot), \mathbb{P}_{\nu_2,S}(\cdot))$:
\begin{equation}
\begin{aligned}
\text{KL}(\mathbb{P}_{\nu_1,S}(\cdot), \mathbb{P}_{\nu_2,S}(\cdot)) = \int_{ X} \mathbb{P}_{\nu_1,S}(X) log\bigg(\frac{\mathbb{P}_{\nu_1,S}(X)}{\mathbb{P}_{\nu_2,S}(X)}\bigg)dX \leq q N \alpha^2.
\end{aligned}
\end{equation}
Here $q$ is a constant via second-order Taylor expansion.

As a consequence, it implies that 
\begin{equation}
\begin{aligned}
 \text{KL}(\mathbb{P}_{\nu_1}, \mathbb{P}_{\nu_2}) \leq q N \alpha^2   \mathbb{E}_{\nu_1}[\N_{S}^t] \leq q N \alpha^2 \frac{\bR_{\nu_1}(t, \pi)}{|\mathcal{U}_{\mathcal{E}}||\Delta_1|}.
\end{aligned}\label{min_max_3}
\end{equation}

The last inequality is due to $S := \arg\min_{S_i\in\U_\bbE} \Delta_1(S_i,S') \mathbb{E}_{\nu_1}[\mathcal{N}_{S_i}^t]$. Combined with Eq~\eqref{min_max_1},~\eqref{min_max_2},~\eqref{min_max_3}:
\begin{equation}
\begin{aligned}
\inf _{\hat{\Delta}_t} \max _{\nu \in \mathcal{E}_0} \mathbb{P}_\nu\left(\max_{i,j \in \U_{\mathcal{E}}}|\hat{\Delta}^{(i,j)}_t-\Delta_\nu^{(i,j)}| \geq \frac{\alpha}{2} \right)  \geq \frac{1}{2}\bigg[1-\sqrt{\frac{1}{2}  q N \alpha^2 \frac{\bR_{\nu_1}(t, \pi)}{|\mathcal{U}_{\mathcal{E}}| |\Delta_{1}|}}\bigg].
\end{aligned}
\end{equation}
 On this basis, we derive the final trade-off as follows:
\begin{equation}
    \begin{aligned}
    &\inf _{\hat{\Delta}_t} \max _{\nu \in \mathcal{E}_0} \mathbb{E}_\nu\left(\max_{i,j \in \U_{\mathcal{E}}}|\hat{\Delta}^{(i,j)}_t-\Delta_\nu^{(i,j)}|\right) \\
    \geq& \frac{\alpha}{2} \inf _{\hat{\Delta}_t} \max _{\nu \in \mathcal{E}_0} \mathbb{P}_\nu\left(\max_{i,j \in \U_{\mathcal{E}}}|\hat{\Delta}^{(i,j)}_t-\Delta_\nu^{(i,j)}| \geq \frac{\alpha}{2} \right) \\ \geq& \frac{\alpha}{4} \bigg[1-\alpha \sqrt{\frac{1}{2} q N \frac{\bR_{\nu_1}(t, \pi)}{|\mathcal{U}_{\mathcal{E}}| |\Delta_{1}|}}\bigg].   \end{aligned}
\end{equation}

% Recall that 
% \begin{equation}
% \begin{aligned}
%  \frac{\alpha}{2}  =& \frac{1}{2N}\sum_{i \in \mathcal{U}} \sum_{A_t \in \mathcal{K}^{\U}} (1-\alpha){f_i(A_t)} \mathbb{P}_{\pi}(A_t|\textbf{S}_t=S ),\\
%  \Delta_{2,S} =& \Delta_{1,S} - 2\frac{\alpha}{2} = \frac{1}{N}\sum_{i \in \mathcal{U}} \sum_{A_t \in \mathcal{K}^{\U}} f_i(A_t) (p_{\pi}(i, A_t,S) - p_{\pi}(i, A_t,S')) - 2\frac{\alpha}{2}.
% \end{aligned}
% \end{equation}

As a consequence, when $t = T$,
\begin{equation}
\begin{aligned}
&\inf _{\hat{\Delta}_T} \max _{\nu \in \mathcal{E}_0} \mathbb{P}_\nu\left(\max_{i,j \in \U_{\mathcal{E}} }|\hat{\Delta}^{(i,j)}_T-\Delta_\nu^{(i,j)}| \geq \frac{\alpha}{2} \right) \sqrt{\bR_{\nu}(T, \pi)} \\
\geq& \inf _{\hat{\Delta}_T} \max _{\nu \in \mathcal{E}_0} 
 {\frac{\alpha}{4} }\bigg[1-\sqrt{\frac{1}{2}q\alpha^2 N \frac{\bR_{\nu_1}(T, \pi)}{|\U_{\mathcal{E}}||\Delta_{1}|}}\bigg] \sqrt{\bR_{\nu_1}(T, \pi)}. \\
%  \geq& 
\end{aligned}\label{final}
\end{equation}

Due to the sqrt-term spans $[0,+\infty]$ with $\alpha \in [0,1]$, hence we could set $q\alpha^2 N \frac{\bR_{\nu_1}(T, \pi)}{|\U_{\mathcal{E}}| |\Delta_{1}|} = \frac{1}{2}$, then, when $T \geq |\mathcal{U}_{\mathcal{E}}|$, it leads to
    \begin{equation}
    \begin{aligned}
    \eqref{final} =& \inf _{\hat{\Delta}_T} \max _{\nu \in \mathcal{E}_0} 
 {\frac{\alpha}{8} }\sqrt{\frac{|\U_{\mathcal{E}}||\Delta_{1}|}{2 N q\alpha^2}} = \Omega_{}(\sqrt{\frac{|\U_{\mathcal{E}}|}{N}}) = \Omega(\sqrt{{|\U_{\mathcal{E}}|}{}}) .
    \end{aligned}
\end{equation}
% When $T \leq |\mathcal{U}_{\mathcal{E}}|$, notice that 
% $\bR_{\nu_2,S}(t, \pi) = O(T) = O(|\mathcal{U}_{\mathcal{E}}|)$, it leads to $\alpha = \Omega(\frac{1}{\sqrt{N}})$, then
% \begin{equation}
% \begin{aligned}
%  \eqref{final} = & \inf _{\hat{\Delta}_t} \max _{\nu \in \mathcal{E}_0}  {\frac{\alpha}{8} }\sqrt{\frac{T}{\sqrt{N}}} = \Omega_{T,N}(\frac{1}{\sqrt{N}}\sqrt{\frac{T}{\sqrt{N}}}) = \Omega_{}({\sqrt{T}}{}).
% \end{aligned}
% \end{equation}

% \zhiheng{$\alpha$ has a non-trivial lower bound, leads to $\sqrt{T}$}.

% \begin{condition}
%     ϕα(t)=Ω(√|log(α)|)\frac{\alpha}{2} = \Omega(\sqrt{|log(\alpha)|}) when α→1\alpha \rightarrow 1.
% \end{condition}
% This condition is natural due to the decreasing order of the log function. 
Theorem~\ref{thm_pareto} also follows. Q.E.D.
    \end{proof}
% \begin{condition}
%     T→+∞T\rightarrow +\infty.
% \end{condition}

\section{Proof of Theorem~\ref{thm_pareto}}\label{app_optimal}
\begin{proof}[Proof of Theorem~\ref{thm_pareto}]
We prove such sufficiency via contradiction. On the one hand, suppose that the MAB pair $\{\pi, \hat{\Delta}\}$ satisfies $\max _{\nu \in \mathcal{E}_0} \Big( \sqrt{\bR_{\nu}(T, \pi)} e_\nu(T, \hat{\Delta}^{}) \Big) = \Tilde{O}(\sqrt{|\mathcal{U}_{\mathcal{E}}|})$. If it is not Pareto-optimal, it is equivalent to claim that there is another pair $\{\pi', \hat{\Delta'}\}$ to dominate $\{\pi, \hat{\Delta}\}$. In this sense, according to Theorem~\ref{trade-off}, there exists an instance $\nu'$ such that $\sqrt{\bR_{\nu'}(T, \pi')} e_{\nu'}(T, \hat{\Delta}^{'})  = \Omega(\sqrt{|\mathcal{U}_{\mathcal{E}}|})$. Moreover, according to the definition of Pareto-dominance, there further exists another instance $\nu''$, such that $\forall \otimes \in \{K, T\}, \sqrt{|\mathcal{U}_{\mathcal{E}}|} \prec_{\otimes}  \sqrt{\bR_{\nu''}(T, \pi)} e_{\nu''}(T, \hat{\Delta}^{})$. It is a contradiction.

\begin{remark}
    On the other hand, we additionally consider the proof of necessity part, also by contradiction. It is a rigorous refinement of Theorem.5 in~\citet{simchi2023multi} with the extension to the network interference case. We additionally condition that 
    $\bR_{\nu}(T, \pi)$ and $e_\nu(T, \hat{\Delta}^{})$ could both be lower bounded by a polynomial form of $T$, i.e., the Pareto-dominance is only considered in the region of $\mathcal{V}_{\text{lower}}:=\{\nu: \mathcal{R}_{\nu}(T,\pi)=\Omega(T^{\alpha}), e_{\nu}(T, \hat{\Delta}^{}) = \Omega(\sqrt{|\mathcal{U}_{\mathcal{E}}|}T^{\beta})\}$, where $\alpha>0, \beta<0$ are constants. Recalling our goal is to prove any Pareto-optimal pair $\{\pi, \hat{\Delta}\}$ satisfies $$\max _{\nu \in \mathcal{V}_{lower}} \Big( \sqrt{\bR_{\nu}(T, \pi)} e_\nu(T, \hat{\Delta}^{}) \Big) = \Tilde{O}\Big(\sqrt{|\mathcal{U}_{\mathcal{E}}|}\Big).$$ Suppose that for a Pareto-optimal pair, there exist hard instances $\nu^* \in \mathcal{V}_{\text{hard}} \subseteq \mathcal{V}_{front}\cap \mathcal{V}_{lower} \subseteq \mathcal{E}_0$ such that (here $\mathcal{V}_{front}:= \{\nu: (\sqrt{\bR_{\nu}(T, \pi)}, e_{\nu}(T, \hat{\Delta}^{})) \in \mathcal{F}(\pi,\hat{\Delta})\}$): $$\forall \nu^* \in \mathcal{V}_{\text{hard}}, \sqrt{\bR_{\nu^*}(T, \pi)} e_{\nu^*}(T, \hat{\Delta}^{}) > C\sqrt{|\mathcal{U}_{\mathcal{E}}|}, \text{when~} T \text{~is sufficiently large}.$$
Here, $C$ is a constant. According to our condition, it induces that $\bR_{\nu}(T, \pi) \succ_T C_1 T^{2\alpha_1}$, $e_\nu(T,\hat{\Delta}) \succ_T C_2 |\mathcal{U}_{\mathcal{E}}|^{1/2}T^{\alpha_2}$, where $C_1,C_2\geq 0, C_1 C_2 = C, \alpha_1+\alpha_2 > 0, \alpha_2 \leq 0, \alpha_1 \in [0,1/2]$ since the regret is bounded as $O(T)$. It indicates that $\alpha_2 \geq -1/2$. On this basis, we could construct feasible pair $\{\pi_{\text{alg}}, \hat{\Delta}_{\text{alg}}\}$ via selecting suitable $T_1 := T^{-2\alpha_2}$ in Algorithm~\ref{alg1} to satisfy $e_\nu(T,\hat{\Delta}) \simeq_T e_\nu(T,\hat{\Delta})$\footnote{Here $\simeq$ is the combination of $\succ$ and $\prec$.}.   According to Theorem~\ref{trade-off1}, it follows that the pair $\{\pi_{\text{alg}}, \hat{\Delta}_{\text{alg}}\}$ would Pareto-dominate the original $\{\pi_{\text{}}, \hat{\Delta}_{\text{}}\}$. Contradiction.

% Then, we could construct compatible pair $\{\pi_{\text{alg}}, \hat{\Delta}_{\text{alg}}\}$ via selecting feasible $T_1$ in Algorithm~\ref{alg1} to make sure $e_{\nu^*}(T, \hat{\Delta}^{}) = \Theta(e_{\nu^*}(T, \hat{\Delta}_{\text{alg}}))$. According to Theorem~\ref{trade-off1}, it induces that ${\bR_{\nu^*}(T, \pi_{\text{alg}})} \prec_{\otimes} {\bR_{\nu^*}(T, \pi)}$, $\forall \otimes \{K,T\}$. For the complementary part, namely, for the case $\nu \in (\mathcal{V}_{lower} \cap \mathcal{V}_{front})/\mathcal{V}_{hard}$, then due to the fact $\sqrt{\bR_{\nu}(T, \pi)} e_\nu(T, \hat{\Delta}^{}) = \Tilde{O}\Big(\sqrt{|\mathcal{U}_{\mathcal{E}}|}\Big)$ and $e_{\nu}(T, \hat{\Delta}^{}) >_{} \Omega(\sqrt{|\mathcal{U}_{\mathcal{E}}|}/\sqrt{T})$, we could also find legitimate construction of $T_1$ in Algorithm~\ref{alg1}. In combination, 
\end{remark}

% Since the regret $\bR_{\nu^*}(T, \pi)$ is non-decreasing corresponding to $T$, we claim there exists a constant $C'$ such that $\bR_{\nu^*}(T, \pi) \geq T^{C'}$. Then 

\end{proof}

\section{Proof of Theorems in Section~\ref{alg}}\label{app_alg}

\subsection{Proof of Theorem \ref{stochasticATE}}

\begin{proof}[Proof of Theorem \ref{stochasticATE}]
Based on the design of the Algorithm \ref{alg1}, in the first phase, we have $ \N_{S}^{T_1} \ge \lfloor \frac{T_1}{\vert \U_\bbE \vert} \rfloor \ge 1$ for all $S \in \U_\bbE$. Define the good event as $\bbE_{T_1} : = \bigg\{  \hat{R}_{T_1}(S) -  \frac{1}{N}\sum_{i\in\U} \tilde{Y}_i(S) \leq \sqrt{{4\log(T_1\vert \U_\bbE \vert)}/{\N_S^{T_1}}},\ \forall S \in \U_\bbE \bigg\}$ and its complement as $\bbE_{T_1}^c$. Based on the previous discussion, the sub-Gaussian proxy of any exposure super arm's reward distribution is at most $2$, then
based on the Hoffeding inequality (Lemma \ref{hoffeding}), we have for a exposure super arm $S\in\U_\bbE$:
    \begin{align}\label{eq49}
\bP \left( \hat{R}_t(S) - \frac{1}{N}\sum_{i \in \U} \tilde{Y}_i(S) > a \right) \leq e^{-\frac{\N_S^t a^2}{4}},
\end{align}
   substituting $t = T_1$ and $a = \sqrt{\frac{4\log(T_1 \vert \U_\bbE \vert)}{\N_S^{T_1}}}$ into Eq~\eqref{eq49} and we can derive
   \begin{align}
\bP \left( \hat{R}_{T_1}(S) - \frac{1}{N}\sum_{i \in \U} \tilde{Y}_i(S) > \sqrt{\frac{4\log\big(T_1\vert \U_\bbE \vert\big)}{\N_S^{T_1}}} \right) \leq  \frac{1}{T_1 \vert \U_\bbE \vert}.
\end{align}
Utilize the union bound, there is
\begin{align}
\begin{split}
    \bP \big( \bbE^c_{T_1} \big)
    \le & \sum_{S\in\U_\bbE} \bP \Bigg(\Bigg\{  \hat{R}_{T_1}(S) - \frac{1}{N}\sum_{i\in\U} \tilde{Y}_i(S) > \sqrt{\frac{4\log\big(T_1\vert \U_\bbE \vert\big)}{\N_S^t}} \Bigg\} \Bigg)\\ \le & \sum_{S\in\U_\bbE} \frac{1}{T_1 \vert \U_\bbE \vert}
    \\
    \le & \frac{1}{T_1},
\end{split}
\end{align}
and $\bP(\bbE_{T_1}) \ge 1 - \frac{1}{T_1}$. Therefore, for all $S_i,S_j \in \U_\bbE$, we have:
    \begin{align}
    \begin{split}
        &\bE \Big[ \big\vert \Delta^{(i,j)} - \hat{\Delta}_T^{(i,j)} \big\vert \Big]\\
        \le & \bP(\bbE_{T_1}) \bE \Big[ \big\vert \Delta^{(i,j)} - \hat{\Delta}_T^{(i,j)} \big\vert \mid \bbE_{T_1} \Big] + \bP(\bbE^c_{T_1}) \bE \Big[ \big\vert \Delta^{(i,j)} - \hat{\Delta}_T^{(i,j)} \big\vert \mid \bbE^c_{T_1} \Big]\\
        \le & \bP(\bbE_{T_1}) \bE \Bigg[ \bigg \vert \hat{R}_t(S) - \frac{1}{N}\sum_{i' \in \U} \tilde{Y}_{i'}(S_i) \bigg\vert + \bigg\vert \hat{R}_t(S) - \frac{1}{N}\sum_{i' \in \U} \tilde{Y}_{i'}(S_j)\bigg\vert \mid \bbE_{T_1} \Bigg] + \frac{1}{T_1} \\
        \le & 2\sqrt{\frac{4 \log\big(T_1 \vert \U_\bbE \vert \big)}{\lfloor \frac{T_1}{\vert \U_\bbE \vert} \rfloor}} + \frac{1}{T_1}
        \\ = & \tilde{O}\Bigg(\sqrt{\frac{\vert \U_\bbE \vert}{T_1}}\Bigg),
    \end{split}
    \end{align}
where the second inequality is owing to the triangle inequality and $\Delta^{(i,j)}$ and $\hat{\Delta}_T^{(i,j)}$ $\in [0,1]$, and the last inequality is owing to $\N_S^{T_1} \ge \lfloor \frac{T_1}{\vert \U_\bbE \vert} \rfloor$. Here we finish the proof of Theorem \ref{stochasticATE}.
\end{proof}

\subsection{Proof of Theorem \ref{stochasticregret}}

In this section, we will first provide an instance-dependent regret upper bound (in the following Lemma \ref{dependent_regret_stochastic}), and then, we will provide an instance-independent regret upper bound based on the instance-dependent one.

\begin{lemma}[Instance-dependent regret] \label{dependent_regret_stochastic} Given any instance that satisfies Condition 1. The regret of the \texttt{UCB-TSN} can be upper bounded as follows
\begin{align}
    \bR(T,\pi) = O \Bigg( \sum_{S_i \not = S^*,\Delta^i >0}\frac{\log\big( T\big)}{\Delta^{i}} + T_1 \Bigg).
\end{align}    
\end{lemma}
\begin{proof}[Proof of Lemma \ref{dependent_regret_stochastic}]
Define $\N^{(t,T)}_S = \sum_{t' = t }^T \bone \{ S_{t'} = S \}$. Besides, define the good event for $S_i$ as:
\begin{align}
    \nonumber
    \bbE_{i} = \Bigg\{ \frac{1}{N}\sum_{i^\prime \in\U}\tilde{Y}_{i^\prime}(S^*) 
    \le \text{UCB}_{t,S^*},\ \forall t\in[T_1 + 1,T] \Bigg\}\cap\Bigg\{ \hat{R}_{\T_i,S_i} + \sqrt{\frac{18\log\big( \frac{1}{\delta} \big)}{\T_i}} \le \frac{1}{N}\sum_{i^\prime\in\U}\tilde{Y}_{i^\prime}(S^*)\Bigg\},
    \end{align}
   where $\T_i = \frac{72\log(1/\delta)}{ (\Delta^i)^2}$ and we utilize $ \hat{R}_{\T_i,S_i}$ to represent $ \hat{R}_{t}(S_i) $ when $ \N^t_{S_i} = \T_i$.
   Based on Lemma \ref{good_bad_event}, we have $\mathbb{P}(\bbE_{i}) \ge 1 - (T - T_1 + 1) \delta$ and its complement has $\mathbb{P}(\bbE_{i}^c) \le (T - T_1 + 1) \delta$.
   
   We can decompose and bound the regret as
\begin{align}\label{regret_decompose}
\begin{split}
    \bR(T,\pi) = & \frac{T}{N} 
 \sum_{i \in \U} 
 \Tilde{Y}_{i}(S^* ) -\mathbb{E}_{\pi} \Bigg[ \sum_{t \in [T]} \sum_{i \in \U} \Tilde{r}_{i,t}(S_t) \Bigg], \\
    \le & \underbrace{\sum_{S_i\not=S^*,\Delta^i > 0} \Delta^i \bE_\pi\Big[ \N_{S_i}^{(T_1+1,T)} \Big]}_{\text{regret in second phase}} + \underbrace{\lceil \frac{T_1}{\U_\bbE} \rceil \sum_{S_i \not = S^*} \Delta^i}_{\text{regret in first phase}} \\
    =& \sum_{S_i\not=S^*,\Delta^i > 0} \bigg( \Delta^i \bE_\pi\Big[ \N_{S_i}^{(T_1 + 1,T)} \mid \bbE_{i} \Big] + \Delta^i \bE_\pi\Big[ \N_{S_i}^{(T_1+1,T)} \mid \bbE_{i}^c \Big] \bigg) + \lceil \frac{T_1}{\U_\bbE} \rceil \sum_{S_i \not = S^*} \Delta^i\\
    \le & \sum_{S_i\not=S^*,\Delta^i > 0} \Delta^i \bE_\pi\Big[ \N_{S_i}^{(T_1 + 1,T)} \mid \bbE_{i} \Big] + T^2\delta + \lceil \frac{T_1}{\U_\bbE} \rceil \sum_{S_i \not = S^*} \Delta^i.
\end{split}
\end{align}
Besides, we want to show that under the event $\bbE_{i}$, we have $\N_{S_i}^{(T_1 + 1,T)} \le \T_i$. 
If $T_1 = T$, then this inequality trivially holds. If $T_1 < T$, suppose $\N_{S_i}^{(T_1 + 1,T)} > \T_i$, then, there exists a time $t_i\in[T_1 + 1,T]$, such that $S_{t_i} = S_i$ ($S_i$ is pulled in round $t_i$), and $\N_{S_i}^{(t_i,T)} = \T_i + 1$. Based on the exploration strategy in Algorithm \ref{alg1}, we have $\text{UCB}_{t_i-1,S_i} \ge \text{UCB}_{t_i - 1,S^*}$. However, based on the definition of the event $\bbE_{i}$, we have
\begin{align}
\begin{split}
\nonumber
    \text{UCB}_{t_i - 1,S^*} &\ge \frac{1}{N}\sum_{i^\prime \in \U}\tilde{Y}_{i^\prime}(S^*) \\ & > \hat{R}_{\T_i,S_i} + \sqrt{\frac{18\log(1/\delta)}{\T_i}} \\& = \hat{R}_{t_i - 1}(S_i) + \sqrt{\frac{18\log(1/\delta)}{\N_{S_i}^{t_i - 1}}} \\ & = \text{UCB}_{t_i-1,S_i},
\end{split}
\end{align}
which contradicts the previous assumption. Therefore, under the event $\bbE_{i}$, we have $\N_{S_k}^T \le \T_k$. Substituting this result and $\delta = 1/T^2$ into Eq~\eqref{regret_decompose}, we have
\begin{align}
    \begin{split}
         \bR(T,\pi) \le & \sum_{S_i\not=S^*,\Delta^i > 0} \Delta^i \bE_\pi \Big[ \N_{S_i}^{(T_1 + 1,T)} \mid \bbE_{i} \Big] + T^2\delta + \lceil \frac{T_1}{\U_\bbE} \rceil \sum_{S_i \not = S^*} \Delta^i\\
         \le & \sum_{S_i \not = S^*,\Delta^i >0} \Delta^i \T_i + 1 + \lceil \frac{T_1}{\U_\bbE} \rceil \sum_{S_i \not = S^*} \Delta^i\\
         \le & \sum_{S_i \not = S^*,\Delta^i >0} \frac{144\log\big(T\big)}{ \Delta^i} +  1 + \lceil \frac{T_1}{\U_\bbE} \rceil \sum_{S_i \not = S^*} \Delta^i\\
          = & O\Bigg( \sum_{S_i \not = S^*,\Delta^i >0} \frac{\log\big(T\big)}{ \Delta^i} + \lceil \frac{T_1}{\U_\bbE} \rceil \sum_{S_i \not = S^*} \Delta^i \Bigg).
    \end{split}
\end{align}
Here we finish the proof of Lemma \ref{dependent_regret_stochastic}.
\end{proof}

The proof of Lemma \ref{dependent_regret_stochastic} relies on the following Lemma \ref{good_bad_event}.

\begin{lemma}\label{good_bad_event}
    We have $\mathbb{P}(\bbE_{i}) \ge 1 - (T - T_1 + 1) \delta$ for all $S_i$ satisfies $S_i \not = S^*$ and $\Delta^i > 0$.
\end{lemma}

\begin{proof}[Proof of Lemma \ref{good_bad_event}] Define the complement of $\bbE_{i}$ as 
\begin{align}
\nonumber
\bbE_{i}^c = \Bigg\{ \frac{1}{N}\sum_{i^\prime\in\U}\tilde{Y}_{i^\prime}(S^*) 
        > \text{UCB}^*_{t},\ \exists t\in[T_1 + 1,T] \Bigg\}\cup\Bigg\{ \hat{R}_{\T_i,S_i} + \sqrt{\frac{18\log\big( \frac{1}{\delta} \big)}{\T_i}} > \frac{1}{N}\sum_{i^\prime\in\U}\tilde{Y}_{i^\prime}(S^*)\Bigg\}.
\end{align}
Based on the union bound, we have
\begin{align}\label{eq39}
\begin{split}
    \bP\big(\bbE_{i}^c\big) \le &\bP\Bigg( \Bigg\{ \frac{1}{N}\sum_{i^\prime\in\U}\tilde{Y}_{i^\prime}(S^*) 
    \ge \text{UCB}_{t,S^*},\ \exists t\in[T_1 + 1,T] \Bigg\}\Bigg)\\
    & + \bP\Bigg(  \Bigg\{ \hat{R}_{\T_i,S_i} + \sqrt{\frac{18\log\big( \frac{1}{\delta} \big)}{\T_i}} \ge \frac{1}{N}\sum_{i^\prime\in\U}\tilde{Y}_{i^\prime}(S^*)\Bigg\} \Bigg)\\
    \le & \sum_{t = T_1 + 1}^T \bP\Bigg( \Bigg\{ \frac{1}{N}\sum_{i^\prime\in\U}\tilde{Y}_{i^\prime}(S^*) 
    \ge \text{UCB}_{t,S^*} \Bigg\}\Bigg)\\
    & + \bP\Bigg(  \Bigg\{ \hat{R}_{\T_i,S_i} + \sqrt{\frac{18\log\big( \frac{1}{\delta} \big)}{\T_i}} \ge \frac{1}{N}\sum_{i^\prime\in\U}\tilde{Y}_{i^\prime}(S^*)\Bigg\} \Bigg).\\
\end{split}
\end{align}
Based on Hoeffding's inequality, we can bound the first term in Eq~\eqref{eq39} by:
\begin{align}\label{eq52}
    \sum_{t = T_1 + 1}^T \bP\Bigg( \Bigg\{ \frac{1}{N}\sum_{i^\prime\in\U}\tilde{Y}_{i^\prime}(S^*) 
    \ge \text{UCB}_{t,S^*} \Bigg\}\Bigg) \le (T - T_1) \delta.
\end{align}
Besides, we can bound the second term in Eq~\eqref{eq39} by:
\begin{align}\label{eq53}
\begin{split}
    &\bP\Bigg(  \Bigg\{ \hat{R}_{\T_i,S_i} + \sqrt{\frac{18\log\big( \frac{1}{\delta} \big)}{\T_i}} \ge \frac{1}{N}\sum_{i^\prime\in\U}\tilde{Y}_{i^\prime}(S^*)\Bigg\} \Bigg)\\
    = & \bP\Bigg(  \Bigg\{ \hat{R}_{\T_i,S_i} - \frac{1}{N}\sum_{i^\prime\in\U}\tilde{Y}_{i^\prime}(S_i) \ge \Delta^i - \sqrt{\frac{18\log\big( \frac{1}{\delta} \big)}{\T_i}}\Bigg\} \Bigg)\\
    \le & \bP\Bigg(  \Bigg\{ \hat{R}_{\T_i,S_i} - \frac{1}{N}\sum_{i^\prime\in\U}\tilde{Y}_{i^\prime}(S_i) \ge \frac{1}{2}\Delta^i \Bigg\} \Bigg)\\
    \le & \text{exp}\bigg( -\frac{\T_i  (\Delta^i)^2}{16} \bigg)\\
    \le & \delta,\\
\end{split}
\end{align}
where the first and last inequality is owing to the definition of $\T_i$, and the second inequality is owing to Hoeffding's inequality. Based on Eq~\eqref{eq52} and Eq~\eqref{eq53}, we have $\bP(\bbE_i) \ge 1 - (T - T_1 + 1)\delta$ for all $S_i$ satisfies $S_i \not = S^*$ and $\Delta^i > 0$. Here we finish the proof of Lemma \ref{good_bad_event}.
\end{proof}

Now we can prove Theorem \ref{stochasticregret}.

\begin{proof}[Proof of Theorem \ref{stochasticregret}]
    In the Proof of Lemma \ref{dependent_regret_stochastic}, we shows that for all $S_i \not = S^*,\ \Delta^i >0$, we have
    \begin{align}
    \bE_\pi\Big[\N_{S_i}^{(T_1 + 1,T)}\Big] \le   \frac{144\log(T)}{{(\Delta^i)}^2} + 1.
    \end{align}
    Define $\Lambda = 6\sqrt{\frac{\vert \U_\bbE \vert\log(T) }{T}}$, we can decompose the regret as
     \begin{align}
     \begin{split}
         \bR(T,\pi) &\le \sum_{S_i\not=S^*,\Delta^i < \Lambda} \Delta^i\bE_\pi\Big[\N_{S_i}^{(T_1 + 1,T)}\Big] + \sum_{S_i\not=S^*,\Delta^i \ge \Lambda} \Delta^i \bE_\pi\Big[\N_{S_i}^{(T_1 + 1,T)}\Big] + \lceil \frac{T_1}{\U_\bbE} \rceil \sum_{S_i \not = S^*} \Delta^i\\
         &\le T \Lambda + \sum_{S_i\not=S^*,\Delta^i\ge\Lambda} \bigg(\frac{144\log(T)}{\Delta^i} + \Delta^i \bigg) + \lceil \frac{T_1}{\U_\bbE} \rceil \sum_{S_i \not = S^*} \Delta^i,\\
         &\le T \Lambda + \frac{144 \vert \U_\bbE \vert \log(T)}{\Lambda} + \bigg(1 + \lceil \frac{T_1}{\U_\bbE} \rceil \bigg) \sum_{S_i \not = S^*} \Delta^i\\
         &\le 30\sqrt{\vert \U_\bbE \vert T \log(T)}  + \bigg(1 + \lceil \frac{T_1}{\U_\bbE} \rceil \bigg) \sum_{S_i \not = S^*} \Delta^i\\
         & = \tilde{O}\Bigg(\sqrt{\vert \U_\bbE \vert T} + \frac{T_1}{\vert \U_\bbE \vert} \sum_{S_i\not = S^*} \Delta^i \Bigg).
     \end{split}
     \end{align}
     Here we finish the proof of Theorem~\ref{stochasticregret}.
\end{proof}

\section{Algorithm for Adversarial Setting in~\citet{simchi2023multi}}\label{app_additional_alg}

\begin{algorithm*}[t]
\renewcommand{\algorithmicrequire}{\textbf{Input:}}
\renewcommand{\algorithmicensure}{\textbf{Output:}}
	\caption{EXP3-Two Stage Network (\texttt{EXP3-TSN})}
	\begin{algorithmic}
         \STATE \textbf{Input:} arm set $\A$, unit number $N$, exposure super arm set $\U_\mathcal{E}$, estimator set $\{\hat{R}_0(S) = 0\}_{S \in \U_\bC}$, active super exposure arm set $\A_0 = \U_\mathcal{E}$, $T_1$, $\alpha = (e-2)(1 +  2\vert\U_\mathcal{E}\vert) e^2\log(2/\delta)$, $\epsilon = \sqrt{\frac{\log(\vert \U_\bbE \vert)}{\vert \U_\bbE \vert T}}$
        \FOR{$t = 1 : T_1$}
        \STATE $\forall S \in \U_\mathcal{E}:$ $\pi_t(S) = \frac{1}{\vert \U_\mathcal{E} \vert}$ and sample $S_t$ based on $\pi_t$
        \STATE Sample $S_t$ based on $\pi_t$, implement \texttt{Sampling}($S_t$)
        \ENDFOR
         \STATE Output $\hat{\Delta}^{(i,j)} = \frac{1}{T_1}\hat{R}_{T_1}(S_i) - \frac{1}{T_1}\hat{R}_{T_1}(S_j)$ for any $S_i,S_j \in \U_\mathcal{E}$, $S_i \not = S_j$
        \STATE $\forall S\in \U_\mathcal{E} :$ set $\hat{R}_{T_1}(S) = 0$
        \FOR{$t = T_1 + 1 : T$}
         \STATE $\forall S \in \U_\bbE$: $\pi_t(S) = \frac{\text{exp}(\epsilon\hat{R}_{t-1}(S))}{\sum_{S \in \bbS_t}
        \text{exp}(\epsilon\hat{R}_{t-1}(S))}$
        \STATE Sample $S_t$ based on $\pi_t$, implement \texttt{Sampling}($S_t$)
        \STATE $\forall$ $S\in\U_\bbE$: set $\hat{R}_t(S) = \hat{R}_{t - 1}(S) + 1 - \frac{\bone\{ S_t = S \} \big(1 - \frac{1}{N}\sum_{i\in\U}\tilde{r}_{i,t}(S) \big)}{\pi_t(S)}$
        \ENDFOR
	\end{algorithmic}  
\end{algorithm*}

This section introduces our algorithm, \texttt{EXP3-TSN}, which operates in two distinct phases. In the first phase, the algorithm uniformly samples exposure super arms from the set $\U_\mathcal{E}$. Upon receiving reward feedback, it leverages this data to build unbiased inverse probability weighting (IPW) estimators to estimate the potential outcomes for the super arms. In the second phase, the algorithm applies the EXP3 strategy to minimize regret effectively.

\paragraph{Unbiased estimators for exposure mapping} We construct unbiased inverse probability weighting (IPW) estimators to estimate the potential outcome of each exposure super arm, i.e.,
\begin{align}
    \hat{R}_t(S) = \hat{R}_{t - 1}(S) + 1 - \frac{\bone\{ S_t = S \} \big(1 - \frac{1}{N}\sum_{i\in\U}\tilde{r}_{i,t}(S) \big)}{\pi_t(S)}.
\end{align}
It is easy to verify that for all $S\in\U_\bbE$, for all $t\in[1,T]$:
\begin{align}
   \bE\Bigg[ 1 - \frac{\bone\{ S_t = S \} \big(1 - \frac{1}{N}\sum_{i\in\U}\tilde{r}_{i,t}(S) \big)}{\pi_t(S)} \mid \bH_{t-1} \Bigg] = \frac{1}{N}\sum_{i\in\U}\tilde{Y}_i(S) + f_t.
\end{align}
Using our unbiased estimator $\hat{R}_{t}(S)$, we can accurately estimate the ATE (which is demonstrated in Theorem \ref{lemmaATE}). We define the martingale sequence as $\big(\{M^{(i,j)}_{t^\prime}\}_{S_i \neq S_j}\big)_{t^\prime = 1}^t$, where 
$M_t^{(i,j)} = \hat{R}_t(S_i) - \hat{R}_t(S_j) - \Delta^{(i,j)}$, and it is easy to verify that $\bE\big[M_t^{(i,j)} \mid \bH_{t-1}\big] = 0$.

% \zhiheng{for any exposure mapping, cluster, condition $1$,  for any instance...for our algorithm (exists one $\{\pi, \hat{\Delta}\}$),...we get the upper bound...In contrast, the ``general'' lower bound...}

% \zhiheng{Simple algorithm with √T\sqrt{T} regret}
% \zhiheng{add the instance-independent algorithm when Δ\Delta is small or KK is large (two equal optimal arm, match su jia). The same analysis with Chonghuan; he can get two regrets.Chonghuan; he can get two regrets. Then we consider a better regret upon KK.}

\section{Proof of Theorem~\ref{thm_adversarial}}\label{app_adver}

Theorem~\ref{thm_adversarial} could be equivalently separated as the following Theorem \ref{lemmaATE} and Theorem~\ref{independent_regret_EXP3}.

\subsection{Proof of Theorem \ref{lemmaATE}}

\begin{theorem}[Bounding the ATE estimation] \label{lemmaATE} Given any instance that satisfy $T \ge \mathcal{T}(T)$ and $\vert \U_\bbE \vert \ge 2$. Set $T \ge T_1 \ge \mathcal{T}(T_1)$. For any $S_i \not = S_j$, the ATE estimation error of the \texttt{EXP3-TS} can be upper bounded as follows: $\bE \Big[ \vert \hat{\Delta}^{(i,j)}_T - \Delta^{(i,j)} \vert \Big] = \tilde{O}\Big(\sqrt{\frac{\vert \U_\mathcal{E} \vert}{T_1}} \Big)$.
\end{theorem}

\begin{proof}[Proof of Theorem \ref{lemmaATE}]
    The proof of this lemma is based on the Bernstein Inequality. To utilize it, we first need to upper bound $\vert M_t^{(i,j)} - M_{t-1}^{(i,j)}\vert$, $\forall t \in [T_1]$. It can be expressed as:
    \begin{align}
    \begin{split}
    \nonumber
        &\big\vert M_t^{(i,j)} - M_{t-1}^{(i,j)} \big\vert \\=& \Bigg\vert \frac{\bone\{S_t = S_i\} \big(1 - \frac{1}{N}\sum_{i^\prime \in \U}\tilde{r}_{i^\prime,t}(S_i)\big)}{\pi_t(S_i)} - \frac{\bone\{S_t = S_j\} \big(1 - \frac{1}{N}\sum_{i^\prime \in \U}\tilde{r}_{i^\prime,t}(S_j)\big)}{\pi_t(S_j)} - \Delta^{(j,i)} \Bigg\vert
        \\
        \le& \frac{1}{\pi_t(S_i)} + \frac{1}{\pi_t(S_j)} + 1\\
        = & 2\vert \U_\mathcal{E} \vert + 1,
    \end{split}
    \end{align}
    where the first inequality is owing to the $\tilde{r}_{i,t}(\cdot)\in[0,1]$ and $\Delta^{(j,i)} \in [-1,1]$, and
    the second equality is due to the definition of $\pi_t(S)$ in the first phase. We also need to upper bound the variance of the martingale in the first phase, denoted as $V_t^{(i,j)}$, i.e.,
\begin{align}
\begin{split}
\nonumber
    &V_t^{(i,j)} \\= &\sum_{t \in [T_1]} \bE \Bigg[ \Bigg( \frac{\bone\{S_t = S_i\} \big(1 - \frac{1}{N}\sum_{i^\prime \in \U}\tilde{r}_{i^\prime,t}(S_i)\big)}{\pi_t(S_i)} - \frac{\bone\{S_t = S_j\} \big(1 - \frac{1}{N}\sum_{i^\prime \in \U}\tilde{r}_{i^\prime,t}(S_j)\big)}{\pi_t(S_j)} - \Delta^{(i,j)} \Bigg)^2 \mid \bH_{t-1} \Bigg]\\
    \le & \sum_{t \in [T_1]} \bigg( \frac{1}{\pi_t(S_i)} + \frac{1}{\pi_t(S_j)} \bigg) \\
    \le & 2 T_1 \vert \U_\mathcal{E} \vert.
\end{split}
\end{align}
Based on this fact that $T_1 \ge \frac{(2\vert \U_\bbE \vert + 1)^2\log(2T_1\vert \U_\bbE \vert^2)}{2(e-2)\vert \U_\bbE \vert}$, we have
\begin{align}
\nonumber
\sqrt{\frac{\log(2T_1\vert \U_\bbE \vert^2)}{2(e-2)\vert \U_\mathcal{E} \vert T_1}} \le \frac{1}{2\vert \U_\mathcal{E} \vert + 1},
\end{align}
which implies we can utilize the Bernstein Inequality (Lemma \ref{berstein}). By the Bernstein inequality, we have: $\forall t \in [T_1]$, with probability at least $1 - \frac{1}{T_1 \vert \U_\mathcal{E} \vert^2}$, there is 
\begin{align}
\nonumber
    \big\vert M^{(i,j)}_t \big\vert \le 2\sqrt{  2(e-2)\vert \U_\mathcal{E} \vert T_1 \log(2T_1\vert \U_\mathcal{E} \vert^2) }.
\end{align}
Dividing both sides by $T_1$, based on the definition of the martingale $M_t^{(i,j)}$ and the ATE estimator $\hat{\Delta}^{(i,j)}$, we have:
\begin{align}
    \big\vert \Delta^{(i,j)} - \hat{\Delta}_T^{(i,j)} \big\vert \le 2\sqrt{ \frac{ 4(e-2)\vert \U_\mathcal{E} \vert\log\big(2T_1 \vert \U_\mathcal{E} \vert\big)}{T_1} }.
\end{align}
Define the good event as $\bbE_{T_1} := \bigg\{\big\vert \Delta^{(i,j)} - \hat{\Delta}_T^{(i,j)} \big\vert \le 2\sqrt{ \frac{ 4(e-2)\vert \U_\mathcal{E} \vert \log(2T_1\vert \U_\mathcal{E} \vert)}{T_1} },\ \forall S_i\not = S_j\bigg\}$. By applying the union bound, it is easy to know that
\begin{align}
    \bP\big( \bbE_{T_1} \big) \ge 1 -  \frac{1}{T_1}.
\end{align}
Based on the above result, for any $S_i \not = S_j$, we have
    \begin{align}
    \begin{split}
        \bE \Big[ \vert \hat{\Delta}_T^{(i,j)} - \Delta^{(i,j)} \vert \Big] \le&
         \bP(\bbE_{T_1}) \bE \Big[ \big\vert \Delta^{(i,j)} - \hat{\Delta}_T^{(i,j)} \big\vert \mid \bbE_{T_1} \Big] + \bP(\bbE^c_{T_1}) \bE \Big[ \big\vert \Delta^{(i,j)} - \hat{\Delta}_T^{(i,j)} \big\vert \mid \bbE^c_{T_1} \Big] \\
        \le & 2\sqrt{ \frac{4 (e-2)\vert \U_\mathcal{E} \vert\log\big(2T_1\vert \U_\mathcal{E} \vert\big)}{T_1} } + \frac{1}{T_1}
        \\ = & \tilde{O}\bigg(\sqrt{\frac{\vert \U_\bbE \vert}{T_1}}\bigg).
    \end{split}
    \end{align}
    Here we finish the proof of Theorem \ref{lemmaATE}.
\end{proof}

\begin{theorem}[Regret upper bound]\label{independent_regret_EXP3}Given any instance that satisfy $T \ge \mathcal{T}(T)$ and $\vert \U_\bbE \vert \ge 2$. The regret of \texttt{EXP3-TS} can be upper bounded by $\bR(T,\pi) = \tilde{O}\big(  \sqrt{\vert \U_\bbE \vert T} + T_1 \big)$.
\end{theorem}

\subsection{Proof of Theorem~\ref{independent_regret_EXP3}}

\begin{proof}[Proof of Theorem \ref{independent_regret_EXP3}]
  Define $R(t,j) = \frac{1}{N} \sum_{i^\prime\in\U} \Big(\tilde{Y}_{i^\prime}(S_j)\Big) + f_t$ as the potential outcome of exposure super arm $S_j\in\U_\bbE$ in round $t$. For all $S_i \in \U_\bbE$, we define 
    \begin{align}\label{eq79}
        \bR(T,\pi,i) = \sum_{t \in [T]} R(t,i) - \bE_\pi \Bigg[ \frac{1}{N} \sum_{t \in [T]} \sum_{i^\prime \in \U}\tilde{r}_{i^\prime,t}(S_t) \Bigg]
    \end{align}
as the expected "regret" if the exposure super arm $S_i$ is the best arm. If we can upper bound $\bR(T,\pi,i)$ for all $S_i\in \U_\bbE$, then we can upper bound $\bR(T,\pi)$. Based on the unbiased property of the IPW estimator, for all $t\in\{T_1 + 1,\dots, T\}$, we have 
\begin{align}\label{eq80}
\begin{split}
&\mathbb{E}_\pi\Big[\hat{R}_{T}(S_i^\prime)\Big] = \sum_{t = T_1 + 1}^T R(t,i^\prime)  \quad \text{and} \\& \mathbb{E}_\pi\Bigg[ \frac{1}{N} \sum_{t \in [T]} \sum_{i^\prime \in \U}\tilde{r}_{i^\prime,t}(S_t) \mid \bH_{t-1} \Bigg] = \sum_{t \in [T]} \sum_{S_{i^\prime}\in\U_\bbE} \pi_t(S_{i^\prime}) R(t,i^\prime) = \sum_{t \in  [T]} \sum_{S_{i^\prime} \in \U_\bbE} \pi_t(S_{i^\prime}) \mathbb{E}_\pi\Big[ \hat{R}_t(S_{i^\prime}) - \hat{R}_{t-1}(S_{i^\prime}) \mid \bH_{t-1}\Big].
\end{split}
\end{align}
Based on Eq~\eqref{eq80}, Eq~\eqref{eq79} can be rewritten as
\begin{align}
\begin{split}
    \bR(T,\pi,i) & \le \bE_\pi [\hat{R}_T(S_i)] - \bE_\pi \Bigg[ \frac{1}{N}\sum_{t = T_1 + 1}^T \sum_{i^\prime\in\U} \tilde{r}_{i^\prime,t}(S_t) \Bigg] + T_1 \\
    & = \bE_\pi [\hat{R}_T(S_i)] - \bE_\pi \Bigg[ \bE_\pi \Bigg[ \frac{1}{N}\sum_{t = T_1 + 1}^T \sum_{i^\prime\in\U} \tilde{r}_{i^\prime,t}(S_t) \mid \bH_{t-1} \Bigg] \Bigg] + T_1 \\
    & = \bE_\pi[\hat{R}_T(S_i)] - \bE_\pi \Bigg[ \sum_{t = T_1 + 1}^T \sum_{S_{i^\prime} \in \U_\bbE} \pi_t(S_{i^\prime})  \bE_\pi \Big[\Big(\hat{R}_t(S_{i^\prime}) - \hat{R}_{t-1}(S_{i^\prime})\Big) \mid \bH_{t-1} \Big] \Bigg] + T_1\\
    & = \bE_\pi \Bigg[ \hat{R}_T(S_i) - \sum_{t = T_1 + 1}^T \sum_{S_{i^\prime} \in \U_\bbE} \pi_t(S_{i^\prime})  \Big(\hat{R}_t(S_{i^\prime}) - \hat{R}_{t-1}(S_{i^\prime})\Big) \Bigg] + T_1\\
    & = \bE_\pi \big[ \hat{R}_T(S_i) - \hat{R}_T \big] + T_1,
\end{split}
\end{align}
where the first and third equality is owing to the tower rule, and the last equality is owing to we define $\hat{R}_T = \sum_{t = T_1 + 1}^T \sum_{S_{i^\prime} \in \U_\bbE} \pi_t(S_{i^\prime})  \Big(\hat{R}_t(S_{i^\prime}) - \hat{R}_{t-1}(S_{i^\prime})\Big)$.

Define $W_T = \sum_{S_{i^\prime}\in\U_\bbE} \text{exp}\big(\epsilon \hat{R}_T(S_{i^\prime})\big)$, we have
\begin{align}\label{eq82}
\begin{split}
    W_T & = W_{T_1}\frac{W_{T_1 + 1}}{W_{T_1}} \cdot\cdot\cdot \frac{W_T}{W_{T-1}} \\ & = \vert \U_\bbE \vert \prod_{t = T_1 + 1}^T \frac{W_{t}}{W_{t - 1}}\\
    & = \vert \U_\bbE \vert \prod_{t = T_1 + 1}^T \Bigg( \sum_{S_{i^\prime} \in \U_\bbE} \frac{\text{exp}\big(\epsilon\hat{R}_{t - 1}(S_{i^\prime})\big)}{W_{t - 1}}\text{exp}\bigg(\epsilon\Big(\hat{R}_{t}(S_{i^\prime}) - \hat{R}_{t - 1}(S_{i^\prime}) \Big)\bigg) \Bigg)\\
    & = \vert \U_\bbE \vert \prod_{t = T_1 + 1}^T \Bigg( \sum_{S_{i^\prime} \in \U_\bbE} \pi_{t}(S_{i^\prime})  \text{exp}\bigg(\epsilon\Big(\hat{R}_{t}(S_{i^\prime}) - \hat{R}_{t - 1}(S_{i^\prime}) \Big)\bigg) \Bigg)\\
     & \le  \vert \U_\bbE \vert \prod_{t = T_1 + 1}^T \Bigg( 1 + \epsilon \sum_{S_{i^\prime} \in \U_\bbE} \pi_{t}(S_{i^\prime})  \Big(\hat{R}_{t}(S_{i^\prime}) - \hat{R}_{t - 1}(S_{i^\prime}) \Big) \\& \quad+ \epsilon^2 \sum_{S_{i^\prime} \in \U_\bbE} \pi_{t}(S_{i^\prime})  \Big(\hat{R}_{t}(S_{i^\prime}) - \hat{R}_{t - 1}(S_{i^\prime}) \Big)^2\Bigg)\\
    & \le \vert \U_\bbE \vert \prod_{t = T_1 + 1}^T \text{exp}\Bigg( \epsilon \sum_{S_{i^\prime} \in \U_\bbE} \pi_{t}(S_{i^\prime})  \Big(\hat{R}_{t}(S_{i^\prime}) - \hat{R}_{t - 1}(S_{i^\prime}) \Big) \\&\quad + \epsilon^2 \sum_{S_{i^\prime} \in \U_\bbE} \pi_{t}(S_{i^\prime})  \Big(\hat{R}_{t}(S_{i^\prime}) - \hat{R}_{t - 1}(S_{i^\prime}) \Big)^2 \Bigg)\\
    &= \vert \U_\bbE \vert \text{exp}\bigg( \epsilon \hat{R}_T + \epsilon^2 \sum_{t' = T_1 + 1}^T\sum_{S_{i^\prime} \in \U_\bbE} \pi_{t}(S_{i^\prime})  \Big(\hat{R}_{t}(S_{i^\prime}) - \hat{R}_{t - 1}(S_{i^\prime}) \Big)^2 \bigg),
\end{split}
\end{align}
where the fourth equality is owing to the definition of $\pi_t(S)$, the first inequality is owing to $\text{exp}(x) \le 1 + x + x^2$ for all $x \le 1$ and $\hat{R}_t(S) - \hat{R}_{t-1}(S) \le 1$ for all exposure super arm $S$, the last inequality is owing to $1 + x \le \text{exp}(x)$ for all $x$, and the last equality is owing to the definition of $\hat{R}_T$. Based on the last term of Eq~\eqref{eq82}, we can derive
\begin{align}
    \hat{R}_T(S_i) - \hat{R}_T \le \frac{\log(\vert \U_\bbE \vert)}{\epsilon} + \epsilon \sum_{t = T_1 + 1}^T\sum_{S_{i^\prime} \in \U_\bbE} \pi_{t}(S_{i^\prime})  \Big(\hat{R}_{t}(S_{i^\prime}) - \hat{R}_{t - 1}(S_{i^\prime}) \Big)^2,
\end{align}
and $\bR(T,\pi,i)$ can be bounded by
\begin{align}
\begin{split}
    \bR(T,\pi,i) & \le \bE_\pi\Big[ \hat{R}_T(S_i) - \hat{R}_T \Big] + T_1\\
    & \le \frac{\log(\vert \U_\bbE \vert)}{\epsilon} + \bE_\pi\Bigg[ \epsilon \sum_{t = T_1 + 1}^T\sum_{S_{i^\prime} \in \U_\bbE} \pi_{t}(S_{i^\prime})  \Big(\hat{R}_{t}(S_{i^\prime}) - \hat{R}_{t - 1}(S_{i^\prime}) \Big)^2 \Bigg] + T_1.
\end{split}
\end{align}
We then try to bound $\bE_\pi\Big[ \epsilon \sum_{t = T_1 + 1}^T\sum_{S_{i^\prime} \in \U_\bbE} \pi_{t}(S_{i^\prime})  \Big(\hat{R}_{t}(S_{i^\prime}) - \hat{R}_{t - 1}(S_{i^\prime}) \Big)^2 \Big]$, define $\tilde{R}(t,j) = 1 - \frac{1}{N} \sum_{i^{\prime}\in\U} \tilde{r}_{i^{\prime},t}(S_{j})$, there is 
\begin{align}
    \begin{split}
    \nonumber
       &\bE_\pi\Bigg[ \epsilon \sum_{t = T_1+1}^T\sum_{S_{i^\prime} \in \U_\bbE} \pi_{t}(S_{i^\prime})  \big(\hat{R}_{t}(S_{i^\prime}) - \hat{R}_{t - 1}(S_{i^\prime}) \big)^2 \Bigg]  \\ = & \bE_\pi\Bigg[ \epsilon \sum_{t = T_1 + 1}^T\sum_{S_{i^\prime} \in \U_\bbE} \pi_{t}(S_{i^\prime}) \bigg( 1 - \frac{\bone\{ S_t = S_{i^\prime}\} \tilde{R}(t,i^\prime) }{\pi_{t}(S_{i^\prime})} \bigg)^2\Bigg]\\
       = & \bE_\pi\Bigg[ \epsilon \sum_{t = T_1 + 1}^T\sum_{S_{i^\prime} \in \U_\bbE} \pi_{t}(S_{i^\prime}) \bigg( 1 - \frac{2 \times \bone\{ S_t = S_{i^\prime}\} \tilde{R}(t,i^\prime) }{\pi_{t}(S_{i^\prime})} + \frac{\bone\{ S_t = S_{i^\prime}\} \big( \tilde{R}(t,i^\prime)\big)^2 }{\pi_{t}(S_{i^\prime})^2} \bigg)\Bigg]\\
       = & \bE_\pi\Bigg[ \epsilon \sum_{t = T_1 + 1}^T \Bigg(  \frac{2}{N} \sum_{i^\prime\in\U} \tilde{r}_{i^\prime,t}(S_{t}) - 1 \Bigg) + \bE_\pi \Bigg[ \epsilon\sum_{t = T_1 + 1}^T\sum_{S_{i^\prime} \in \U_\bbE} \pi_{t}(S_{i^\prime}) \bigg( \frac{\bone\{ S_t = S_{i^\prime}\} \big( \tilde{R}_{t,i^\prime}\big)^2 }{\pi_{t}(S_{i^\prime})^2} \bigg) \mid \bH_{t - 1} \Bigg]\Bigg]\\
       = & \bE_\pi\Bigg[ \epsilon \sum_{t = T_1 + 1}^T \Bigg(  \frac{2}{N} \sum_{i^\prime\in\U} \tilde{r}_{i^\prime,t}(S_{t}) - 1 \Bigg) +  \epsilon\sum_{t = T_1 + 1}^T \sum_{S_{i^\prime} \in \U_\bbE} \big(\tilde{R}_{t,i^\prime}\big)^2  \Bigg]\\
       \le & \vert \U_\bbE \vert T \epsilon.
    \end{split}
\end{align}
Based on the definition of $\epsilon$, we can finally bound $\bR(T,\pi,i)$ by $\sqrt{\vert \U_\bbE \vert T \log(\vert \U_\bbE \vert)} + T_1$. Here we finish the proof of Theorem \ref{independent_regret_EXP3}.
\end{proof}

\section{Optimization perspective}\label{optimize_perspective}

We provide more justification upon Condition~\ref{condition_in}. Notice that we search the best arm within $\mathcal{U}_{\mathcal{E}} = \mathcal{U}_{\mathcal{C}} \cap \mathcal{U}_{\mathcal{O}}$, then a natural question arises that how to search elements of the intersection of these two sets? What if it is an empty set? The optimization problem is formalized as follows:
\begin{equation}
    \begin{aligned}
    &\sum_{i= 1}^C c_i \textbf{e}_i\\
    s.t.~&\forall i \in \mathcal{U}, c_i \in \mathcal{U}_s, \\
        &\exists A \in K^{\mathcal{U}}, d_M \Big(\big(\textbf{S}(i,A,\mathbb{H})\big)_{i \in \mathcal{U}} , \sum_{i= 1}^C c_i \textbf{e}_i \Big) = 0. 
    \end{aligned}\label{original_formulation}
\end{equation}
Here $\textbf{e}_i$ is a binary indicator $\left(\mathbf{e}_i\right)_j= \begin{cases}1, & \text { if } j \in \mathcal{C}_i \\ 0, & \text { if } j \notin \mathcal{C}_i\end{cases}$. Moreover, $d_M$ denotes the Manhattan Distance.

\paragraph{Searching efficiency} It would be an NP-hard problem with a high computation load without additional assumptions. However, we argue that when we select many common exposure mapping structures, the optimization problem may degenerate into a simpler case, such as an integer programming problem. Consider the mapping $\textbf{S}(i,A,\mathbb{H}) := \boldsymbol{S}(i, A, \mathbb{H}) := \bone\{{\sum_{j\in\U} h_{ij} a_{j}>0}\}$. Then Eq~\eqref{original_formulation} could be transformed to 
\begin{equation}
    \begin{aligned}
    &\sum_{i= 1}^C \bone\{{\sum_{j\in\U} h_{ij} a_{j}}>0 \} \textbf{e}_i \\
 s.t.~     &\exists A \in K^{\mathcal{U}}, \forall p,q \text{~satisfying~}\mathcal{C}^{-1}(p) = \mathcal{C}^{-1}(q), \\&
     \bone \big(\sum_{j \in \mathcal{U}}h_{pj}a_j>0\big) = \bone\big(\sum_{j \in \mathcal{U}}h_{qj}a_j>0\big) .
    \end{aligned}
\end{equation}
To solve it, we recommend practitioners adopt the off-the-shelf optimization techniques in Mixed-Integer Nonlinear Programming~\citet{belotti2013mixed}.

\paragraph{Practical issue} Another question arises: what if Condition~\ref{condition_in} fails, even if it is easy to satisfy via adjusting legitimate exposure mapping function and clustering strategy? We formalize it as a relaxed optimization problem and claim its impact on previous modeling is negligible under mild assumptions upon interference effect:
\begin{equation}
    \begin{aligned}
    \forall {\{c_i\}_{i \in [C]}}, \min_{A\in \mathcal{K}^{\mathcal{U}}} d_M \Big(\big(\textbf{S}(i,A,\mathbb{H})\big)_{i \in \mathcal{U}} , \sum_{i= 1}^C c_i \textbf{e}_i \Big).
    \end{aligned}\label{approximation}
\end{equation}
Apparently, when Condition~\ref{condition_in} is violated, then $\max_{\{c_i\}_{i \in [C]}} \min_{A\in \mathcal{K}^{\mathcal{U}}} d_M \Big(\big(\textbf{S}(i,A,\mathbb{H})\big)_{i \in \mathcal{U}} , \sum_{i= 1}^C c_i \textbf{e}_i \Big) >0$. We recommend practitioners to collect the most \textit{similar} exposure arm compared to the form $\sum_{i= 1}^C c_i \textbf{e}_i$ as above to substitute the original intersection set $\mathcal{U}_{\mathcal{E}}$. Specifically, $\forall {\{c_i\}_{i \in [C]}}$, we collect $\{\textbf{S}(i, \textbf{A}', \mathbb{H})\}_{i \in \mathcal{U}}, \text{~where~} \textbf{A}':= \arg\min_{A\in \mathcal{K}^{\mathcal{U}}} d_M \Big(\big(\textbf{S}(i, A, \mathbb{H})\big)_{i \in \mathcal{U}}, \sum_{i= 1}^C c_i \textbf{e}_i \Big)$ as a substitute of the original corresponding cluster-wise super exposure arm. We call the substituted exposure arm set as $\mathcal{U}_{\mathcal{E}}'$.
% When such approximation error is under control, and under mild conditions as below
% \begin{condition}[Approximation error and neighbourhood interference]\label{condition_add}\\
%     (i) $\max_{\{c_i\}_{i \in [C]}} \min_{A\in \mathcal{K}^{\mathcal{U}}} d_M \Big(\big(\textbf{S}(i,A,\mathbb{H})\big)_{i \in \mathcal{U}} , \sum_{i= 1}^C c_i \textbf{e}_i \Big) \leq k_0$;\\ (ii) the neighborhood number is up to $k_1$, where $k_0,k_1$ is a pre-specified integer. Moreover, the outcome is only affected via neighbourhood treatments.
% \end{condition}

In this sense, we recommend practitioners to re-define the arm as (modified from~\eqref{exposure})
\begin{equation}
\begin{aligned}
[\Tilde{Y}^{\text{ideal}}_i({S}_{t}), \Tilde{r}^{\text{ideal}}_{i,t}({S}_{t})]^{\top} := \sum_{A \in \arg\min_{A'\in \mathcal{K}^{\mathcal{U}}} d_M \big((\textbf{S}(i,A',\mathbb{H}))_{i \in \mathcal{U}} , S_t\big) } [Y_i(A), {r}_{i,t}(A)]^{\top} \mathbb{P}_{}(A_t =A \mid {S}_t).
\end{aligned}
\end{equation}

We denote the newly collected \textit{similar} arm of the ideally best arm $S^*$ as $S^{*}_{\text{real}}$, where the former is constructed via cluster-wise exposure arm (might not be compatible with the original arm), and the latter is defined as
\begin{equation}
    S^{*}_{\text{real}} := \textbf{S}(i,A^*_{\text{real}},\mathbb{H}), \text{~where~} A^*_{\text{real}} \in \arg\min_{A'\in \mathcal{K}^{\mathcal{U}}} d_M \big((\textbf{S}(i,A',\mathbb{H}))_{i \in \mathcal{U}} , S^*\big).
\end{equation}

It could be verified that under legitimate policy $\pi$ (such as uniform sampling), it leads to $\tilde{Y}^{\text{ideal}}_i(S^{*}) = \tilde{Y}_i(S^{*}_{\text{real}})$. Furthermore, the remaining part of the regret analysis could be replicated from the main text, paying attention to the new selection set $\mathcal{U}_{\mathcal{E}}'$.

\section{Auxiliary Lemmas}\label{app_auxiliary}

\begin{lemma}[Sub-Gaussian]
A random variable \( X \) is said to be \textbf{sub-Gaussian} if there exists a constant \( \sigma > 0 \) such that for all \( m \in \mathbb{R} \), the moment generating function of \( X \) satisfies:
\[
\mathbb{E}\left[e^{mX}\right] \le e^{\frac{\sigma^2 m^2}{2}}.
\]
The smallest such \( \sigma^2 \) is known as the sub-Gaussian proxy of \( X \). 
\end{lemma}

\begin{lemma}[Hoeffding's Inequality]\label{hoffeding}
Let $X_1, X_2, \cdots, X_n$ i.i.d. drawn from a $\sigma$-sub-Gaussian distribution, $\overline{X} = \frac{1}{n} \sum_{i=1}^n X_i$ and $\mathbb{E}[X]$ be the mean, then we have
\[
\mathbb{P}\left( \overline{X} - \mathbb{E}[X] \geq a \right) \leq e^{-n a^2 / 2 \sigma^2} \quad \text{and} \quad 
\mathbb{P}\left( \overline{X} - \mathbb{E}[X] \leq -a \right) \leq e^{-n a^2 / 2 \sigma^2}.
\]
\end{lemma}

\begin{lemma}[Bernstein's Inequality]\label{berstein}
    Let \( X_1, X_2, \dots , X_n \) be a martingale difference sequence, where each \( X_t \) satisfies \( |X_t| \leq \alpha \) almost surely for a non-decreasing deterministic sequence \( \alpha_1, \alpha_2, \dots , \alpha_n \). Define \( M_t := \sum_{t'=1}^t X_{\tau} \) as the cumulative sum up to time \( t \), forming a martingale. Let \( \overline{V}_1, \overline{V}_2, \dots , \bar{V}_n \) be deterministic upper bounds on the variance $V_t := \sum_{t'=1}^t \mathbb{E}[X_{\tau}^2 | X_1, \dots, X_{t'-1}]$ of the martingale \( M_t \), and suppose \( \overline{V}_t \) satisfies the condition
\[
\sqrt{\frac{\ln\left(\frac{2}{\delta}\right)}{(e - 2) \overline{V}_t}} \leq \frac{1}{\alpha}.
\]
Then, with probability at least \( 1 - \delta \) for all \( t \):
\[
|M_t| \leq 2 \sqrt{(e - 2) \overline{V}_t \ln \frac{2}{\delta}}.
\]
\end{lemma}

\end{document}